\documentclass[journal]{IEEEtran}
\usepackage{xcolor}

\ifCLASSINFOpdf
\else
\fi
\usepackage{pgfplots} 
\pgfplotsset{compat=1.11,
    /pgfplots/ybar legend/.style={
    /pgfplots/legend image code/.code={%
       \draw[##1,/tikz/.cd,yshift=-0.25em]
        (0cm,0cm) rectangle (3pt,0.8em);},
   },
}
\usepackage{tikz}
\usetikzlibrary{shapes,arrows}

\definecolor{BLACK}{rgb}{0,0,0}
\definecolor{WHITE}{rgb}{1,1,1}
\definecolor{RED}{rgb}{1,0.0,0.0}
\definecolor{BLUE2}{rgb}{0.0,0.3,0.6}
\definecolor{DARKSILVER}{rgb}{0.5,0.5,0.5}
\definecolor{CYAN}{rgb}{0.0,1.0,1.0}
\definecolor{GREEN2}{rgb}{0.2,1.0,0.0}
\definecolor{YELLOW}{rgb}{1.0,0.88,0.21}
\definecolor{DARKRED}{rgb}{0.5,0.0,0.13}
\definecolor{LIGHTRED}{rgb}{0.8,0.0,0.0}
\definecolor{LIGHTPURPLE}{rgb}{0.75,0.58,0.89}
\definecolor{PURPLE}{rgb}{0.54,0.17,0.89}
\definecolor{CYAN}{rgb}{0.3,0.91,0.87}
\definecolor{BLUEGREEN}{rgb}{0.11,0.67,0.84}
\definecolor{DARKGREEN}{rgb}{0.0,0.5,0.0}
\definecolor{LIGHTGREEN}{rgb}{0.4,1.0,0.0}
\definecolor{SILVER}{rgb}{0.75,0.75,0.75}
\definecolor{DARKSILVER}{rgb}{0.5,0.5,0.5}
\definecolor{BLUEBROWN}{rgb}{0.52,0.4,0.27}

\usepackage{booktabs}
\usepackage{graphicx}
\usepackage{caption}
\usepackage{url}
\usepackage{multirow}
\usepackage{subcaption}
\usepackage{amsmath}
\usepackage{float}
\floatstyle{plaintop}
\restylefloat{table}
\begin{document}

%
\title{Learning a Joint Embedding of Multiple Satellite Sensors: A Case Study for Lake Ice Monitoring}
%
%
%

\author{Manu~Tom$^*$,
        Yuchang~Jiang,
        Emmanuel~Baltsavias,
        and~Konrad~Schindler,~\IEEEmembership{Senior~Member,~IEEE}
\thanks{The authors are with the Chair of Photogrammetry and Remote Sensing, Swiss Federal Institute of Technology, Zurich, CH-8093, Switzerland. The first author is also affiliated with the Remote Sensing Group, Swiss Federal Institute of Aquatic Science and Technology, Dübendorf, CH-8600, Switzerland, and with the Chair of Glaciology and Geomorphodynamics, University of Zurich, Zurich, CH-8057, Switzerland. E-mails:
manu.tom@eawag.ch; yujiang@student.ethz.ch; emmanuel.baltsavias@geod.baug.ethz.ch; konrad.schindler@geod.baug.ethz.ch}
\thanks{$^*$ Corresponding author, E-mail: manu.tom@eawag.ch, Phone: +41 58 765 6823, Fax: +41 58 765 5802, Address: BU D18, Eawag, Überlandstrasse 133, CH-8600 Dübendorf, Switzerland.}
\thanks{\textbf{© 2022 IEEE. Personal use of this material is permitted. Permission from IEEE must be obtained for all other uses, in any current or future media, including reprinting/republishing this material for advertising or promotional purposes, creating new collective works, for resale or redistribution to servers or lists, or reuse of any copyrighted component of this work in other works. DOI (published version): 10.1109/TGRS.2022.3211184}}

}
\maketitle

\begin{abstract}
Fusing satellite imagery acquired with different sensors has been a long-standing challenge of Earth observation, particularly across different modalities such as optical and Synthetic Aperture Radar (SAR) images. Here, we explore the joint analysis of imagery from different sensors in the light of representation learning: we propose to learn a joint embedding of multiple satellite sensors within a deep neural network. Our application problem is the monitoring of lake ice on Alpine lakes. To reach the temporal resolution requirement of the Swiss Global Climate Observing System (GCOS) office, we combine three image sources: Sentinel-1 SAR (S1-SAR), Terra MODIS and Suomi-NPP VIIRS. The large gaps between the optical and SAR domains and between the sensor resolutions make this a challenging instance of the sensor fusion problem. Our approach can be classified as a late fusion that is learnt in a data-driven manner. The proposed network architecture has separate encoding branches for each image sensor, which feed into a single latent embedding. I.e., a common feature representation shared by all inputs, such that subsequent processing steps deliver comparable output irrespective of which sort of input image was used. By fusing satellite data, we map lake ice at a temporal resolution of \textless1.5 days. The network produces spatially explicit lake ice maps with pixel-wise accuracies \textgreater91\% (respectively, mIoU scores \textgreater60\%) and generalises well across different lakes and winters. Moreover, it sets a new state-of-the-art for determining the important ice-on and ice-off dates for the target lakes, in many cases meeting the GCOS requirement.
\end{abstract}


%
\IEEEpeerreviewmaketitle

\section{Introduction}
Multi-modal satellite data analysis is an important capability and an active area of research in remote sensing and Earth observation~\cite{SchmittZhu_GRSM_2016}. Its aim is to combine the data acquired with different sensors. With the ever-increasing number of operational satellites and the greater variety of imaging sensors, such a combined analysis becomes even more important. One important advantage of combining (or "fusing") data from multiple satellites is a higher temporal sampling frequency, so as to obtain a denser time-series of dynamic processes.
\par
A whole body of literature exists on combining satellite data from different sensors, at various levels of processing (pixel level, feature level, decision level). In most cases, the fused data have a comparable spatial resolution, e.g., S1-SAR and Sentinel-2 (S2), TerraSAR-X and ALOS PRISM, etc. Here we aim to merge several sensors with a high temporal resolution, leading to a large gap in spatial resolution: while S1-SAR acquires an image of our mid-latitude target area every 1.5-4.5 days at $\approx$20m Ground Sampling Distance (GSD), optical sensors with similar revisit times have low spatial resolutions (GSD of 250m for MODIS, respectively $\approx$375m for VIIRS). See Table~\ref{table:satellite_data}.
Due to the large resolution gap we perform a late fusion. This makes it possible to use texture information from the high-resolution S1-SAR and circumvents pixel-accurate matching and co-registration, which is challenging across different modalities with very different radiometries and large changes in GSD, with associated uncertainties in absolute geo-location and relative co-registration.
\par
Our goal here is to analyse a time-series composed of images from different sensors. Intuitively, this problem is solved if we can transform all input images to a common feature space that preserves the information necessary for the intended downstream task, in our case the detection of lake ice.
We leverage the ability of deep neural networks to learn at the same time a complex, non-linear mapping of the inputs into a latent feature representation, via sequences of convolutions and element-wise, non-linear activation functions; and the subsequent mapping from that representation to the desired output variables.
Our proposed network includes a separate encoder branch with individual layout and weights for each input sensor, to map its raw image values to predictive features. However, the different encoders all share the same feature embedding, which forms the input for two branches: an auxiliary branch that outputs a per-pixel classification into frozen (including snow-covered ice and snow-free ice) or non-frozen (water) and background classes, and serves mainly to train the representation; and a branch that regresses the fraction of the lake covered by water (non-frozen pixels).
The joint embedding means that the encoder branches must learn to produce a single representation that is shared across all sensors (respectively, branches), while at the same time being a suitable basis for the prediction tasks.
\par
Fusion of multi-source satellite data has been employed for a wide variety of Earth observation tasks, including land cover classification, object detection, change detection, etc. For instance, SAR-optical fusion has been applied to sharpen (spatially) low-resolution optical images by fusing them with high-resolution SAR images~\cite{Sanli_ISPRS_2013}, to improve the information extraction by exploiting the complementary nature of radar and optical sensing~\cite{SchmittZhu_GRSM_2016}, and for cloud removal from optical images~\cite{Gao_RS_2020,Meraner2020_ISPRSJ}.
In this paper, we use fusion to construct denser time-series from heterogeneous data, for the specific application of lake ice monitoring. To that end, we infer two outputs: a spatially explicit map of frozen and non-frozen lake pixels at every time step; and a time-series of the fraction of water, which serves as a basis for estimating the critical events of lake ice phenology (LIP), such as \textit{ice-on} and \textit{ice-off} dates.
\par
Lake freezing and thawing are strongly correlated to local warming of the atmosphere \cite{Tom2021_RSE_submission,Qinghai2020}, and \textit{Lake Ice Cover} has been included in the list of essential climate variables~\cite{world_meteorological_organization_2019}. 
Hence, lake ice monitoring is important to support climate science and cryosphere research, and ultimately to support climate change mitigation~\cite{rolnick2019tackling}. 
\par 
For lake ice monitoring, the requirements of the Swiss GCOS office are daily observations and accuracy of $\pm2$ days for the critical ice-on/off dates~\cite{LIP1_final_report_2019}. Several papers have discussed lake ice observation with the help of machine learning, using a single satellite sensor \cite{Tom_isprs_2018_optical,tom_isprs_2020_sar,Wu_Duguay_2021_RSE}, but unfortunately, none of those sensors have so far met the criteria of $\pm$2 days. In earlier work~\cite{tom2020_RS_journal} we have integrated MODIS and VIIRS data at the decision level but still failed to meet the GCOS target in most cases.
Terrestrial webcams deliver image sequences that would satisfy the GCOS criteria \cite{muyan_lakeice_2018,prabha_tom_2020}, but unlike satellite-based methods, they are not suitable for systematic and large-scale monitoring, since most lakes do not have webcams observing them, and even if there is a camera its field of view rarely covers the entire lake surface.
\par
In this work, we combine data from multiple satellite sources to improve the temporal resolution, and thus the accuracy of the critical LIP events around the freeze-up and break-up periods.
MODIS and VIIRS satellite images are available daily, at the cost of reduced spatial resolution. Still, the GSD is adequate for all but the smallest lake. The main issue with optical satellite images is data loss due to clouds, which can significantly decrease the Effective Temporal Resolution (ETR). A case in point is our target application of monitoring Alpine lakes, as they are frequently obscured by clouds, especially around the important ice-on/-off dates.
This means that adding further optical sensors, such as S2 or Landsat-8, will not solve the problem; as they are equally affected by clouds. On the contrary, SAR will not be influenced by clouds. S1-SAR provides regular coverage (for the target regions in Switzerland at least every 4.5 days, but in the best case even every 1.5 days, depending on the number of usable orbits as well as data availability in the early stages of the mission). See Table~\ref{table:satellite_data} for details.
Contrary to the moderate-resolution optical sensors, S1-SAR has a much higher spatial resolution. Therefore, texture analysis becomes important, while at the same time the correct texture description for the task is not obvious. A natural solution is to use a deep convolutional network. Note also, the much higher resolution means many more pixels are available within the area of a given lake, playing to the strength of data-hungry deep networks. 
Lake ice mapping with S1-SAR has been shown to work rather well and has the added advantage that even tiny lakes can be monitored, but by itself, the revisit time is not sufficient to meet the GCOS specification~\cite{tom_isprs_2020_sar}. Below we will show that, at least for our target lakes, this target can in many cases be met with the combination of MODIS, VIIRS and S1-SAR.
\begin{table*}[th!]
\centering
\resizebox{0.8\textwidth}{!}{%
\begin{tabular}{llll}
 \toprule
& \textbf{MODIS} & \textbf{VIIRS} & \textbf{Sentinel-1 SAR}\\
 \midrule
\textit{Satellite type} & optical & optical & radar \\
\textit{Spatial resolution} & 250--1000m    & 375--750m & $\approx$20m\\
\textit{Temporal resolution} & 1d & 1d  & 1.5--4.5d (for Switzerland)\\
\textit{Spectral resolution} & 36 bands & 22 bands & C-band (3.8--7.5cm), \\
 & (0.4--14.2$\mu m$) & (0.4--12.0$\mu m$) & 4 polarisations (mainly VV, VH)\\
\textit{Cloud problems} & severe & severe & nil\\
\textit{Cloud mask issues} & slight & slight & NA\\
\textit{Costs} & free & free & free\\
\textit{Availability} & very good & very good & very good\\
 & (via VIIRS continuity) &  & (HV / HH only partially available)\\
\bottomrule
\end{tabular}}
\caption{Details about the satellite data that we use.}
\label{table:satellite_data}
\end{table*} 


\subsection{Related work}
Satellite data fusion has been a long-standing topic in remote sensing and beyond. Arguably one of the most basic forms of data fusion is pan-sharpening, i.e., fusing an image with high spectral, but low spatial resolution with one that has high spatial resolution but only a single (normally broader) spectral band, eg., Ehlers et al.~\cite{Ehlers_IJIDF_2010}. Related are more complicated spectral-spatial fusion scenarios \cite{Ranchin_PERS_2000,Melgani_PRL_2002,Huang_TGRS_2012,Huang_TGRS_2013,Huang_RSL_2013,Lanaras_RS_2017,Lanaras_CVPRW_2017,Lanaras_PRS_2018} as well as spatio-temporal fusion techniques \cite{song2018spatiotemporal,zhu2018spatiotemporal}.
For readers looking for a general overview of remote sensing data fusion, there are several well-crafted review papers~\cite{Zhang_review_2010,Joshi_RS_review_2016,SchmittZhu_GRSM_2016, Schmitt_IGARSS_2017,Samadhan2020_review_Journal} that discuss technical challenges, solutions, applications, and trends.  
\par
Here, we will limit our review to directly related work, by which we mean recent (deep) representation learning methods in the context of SAR-optical satellite data fusion. 
Furthermore, on the application side, we will review learning-based approaches to lake ice monitoring.

\subsubsection{Fusing optical and SAR satellite images with deep learning}
Several authors have used Convolutional Neural Networks (CNNs) to integrate the complementary information in optical and SAR data, typically using the concept of two-stream networks with two input branches (one for each modality). Mou et al.~\cite{Lichao_2017} proposed to use a two-stream pseudo-Siamese CNN ("SARptical Convolutional Network") to match patches of urban scenes in very high spatial resolution TerraSAR-X (1.25m pixel spacing) and airborne UltraCAM optical imagery (20 cm GSD). They achieved an overall accuracy of 97.48\% at a 0.05\% false alarm rate, on a relatively small dataset with 109 SAR and 9 optical images. Merkle et al.~\cite{Merkle_RS_2017} put forward another two-stream, Siamese CNN to extract features from both optical (PRISM, 2.5m GSD) and SAR (TerraSAR-X) images, followed by a dot-product layer to compute similarities between the extracted features. Their primary goal was to improve the geo-location of optical images, by precisely co-registering them with the corresponding SAR data.
Scarpa et al.~\cite{Scarpa_RS_2018} used SAR imagery (S1-SAR), in conjunction with CNN-based data fusion, to estimate spectral features for cloudy days where optical data (S2) is unusable. 
\par
A dataset "SEN1-2" with image pairs from S1-SAR (VV polarisation) and S2 (only RGB channels) has been proposed by Schmitt et al.~\cite{Schmitt_Hughes_2018} to support research into geo-spatial data fusion. As a follow-up, another curated dataset "SEN12MS" of geo-coded multi-spectral satellite imagery was also made available, which includes patch triplets (dual-polarised S1-SAR, multi-spectral S2 and MODIS land cover) custom-tailored to the training of deep learning methods~\cite{Schmitt_Hughes_2019}.
In another recent work~\cite{Buergmann2019deepfusion} ground control points are derived from high spatial resolution TerraSAR-X imagery (1.25m pixel spacing) to improve the absolute geo-location of optical images (Pl{\'e}iades, 0.5m and 2m GSD for panchromatic and multi-spectral bands, respectively). That method used an adapted version of HardNet~\cite{Mishchuk_2017} pre-trained on the SEN1-2 dataset, which was then fine-tuned on the high spatial resolution target data.
Hoffmann et al.~\cite{Hoffmann_2019} proposed a fully convolutional neural network to predict the similarity metric between SAR and optical images, and reported large improvements over standard metrics based on mutual information, for a subset of the SEN1-2 data.
Wang et al.~\cite{Wang2018_PRS_Journal} performed registration of Landsat, Radarsat and SPOT imagery, also with a deep network. They explored transfer learning to save training time. Additionally, they utilised a self-learning trick to work around the lack of labelled training data. Still concerning SAR-to-optical registration, Hughes et al.~\cite{hughes2020deep} designed a three-step framework, consisting of a "goodness" network that localised image regions suitable for matching, followed by a correspondence network that generated a heatmap of matching scores, and a subsequent outlier rejection network.

Finally, there have also been some studies about the potential of Generative Adversarial Networks (GANs). Hughes et al.~\cite{Hughes_RS_2018} matched TerraSAR-X and UltraCAM data. To lower the rate of false-positive matches, they mined hard negative samples and trained a variational autoencoder (VAE) with an adversarial strategy, to learn the latent distribution of the training data and synthesised realistic "negative" patches.
GANs were also adopted to remove clouds from optical satellite images~\cite{Gao_RS_2020}. In a first step, a CNN estimates an optical image from the cloud-free SAR image ("image-to-image translation"). In the next step, that synthesised image is fused with the original, cloudy optical image to replace the cloud pixels, using a GAN. The method was demonstrated on three different sensor pairs, namely S1/S2, Gaofen-3/2, and airborne SAR/optical images. 

\subsubsection{Lake ice monitoring}
The literature on ice monitoring is vast, as a diverse range of data sources and models have been investigated. Here, we focus on research that used machine learning to derive the relationship between image observations and the presence of ice. In our own earlier work~\cite{Tom_isprs_2018_optical} we have demonstrated semantic segmentation of MODIS and VIIRS data into frozen and non-frozen pixels, using Support Vector Machines~\cite{SVM} on the raw channel intensities. The two sensors were then combined via decision-level fusion to extract the ice-on/-off dates for two winters~\cite{tom2020_RS_journal}. In many cases, the critical dates were determined at an accuracy of 1-4 days. We also confirmed that, for the same target lakes, a deep learning model (Deeplab V3+~\cite{deeplab}) could perform the frozen/non-frozen segmentation of S1-SAR amplitude images, but still failed to determine the ice-on/-off dates with sufficient accuracy~\cite{tom_isprs_2020_sar}.
\par
Hoekstra et al.~\cite{Hoekstra2020} combined Iterative Region Growing Segmentation (IRGS) with a Random Forest (RF) classifier~\cite{randomforest} to distinguish lake ice from open water in RADARSAT-2 images of Great Bear Lake (Canada) from 2013 to 2016, attaining an overall segmentation accuracy of $\approx$96\%. %
Wu et al.~\cite{Wu_Duguay_2021_RSE} have recently compared the performance of four popular machine learning classifiers (SVM, RF, Gradient Boosted Trees [GBT], Multinomial Logistic Regression) for lake ice detection from MODIS, for 17 large lakes with areas \textgreater1040 km\textsuperscript{2}, situated across Northern Hemisphere. They reported the best performance with RF and GBT.
For our small Alpine lakes, we have also performed a similar comparison with SVM, RF and XGBoost~\cite{xgboost}, but found that a linear SVM achieved the best generalisation across different lakes and years~\cite{Tom_thesis_2021}, due to non-linear methods over-fitting to the comparatively small number of training pixels.
\par
Beyond satellite images, there have been attempts to use webcam streams or even crowd-sourced amateur photographs for lake ice monitoring~\cite{muyan_lakeice_2018,prabha_tom_2020}.
To that end, the state-of-the-art encoder-decoder network for semantic segmentation of close-range images~\cite{Jegou2016_CVPRW} was retrained (with minor modifications) to distinguish the relevant classes such as ice, snow and water.
While ice detection based on webcam images works rather well and largely is unaffected by cloud cover (although a few images may have to be dropped because of excessive fog, rain or snowfall), it is at this point unclear how to scale up such an approach, as only a small portion of all relevant lakes are (often only partially) observed by webcams.

\subsection{Definitions used}
Following Franssen and Scherrer~\cite{franssen2008freezing}, we define \textit{ice-on} as the first date on which a lake is "almost fully" frozen, and followed by a second day where this remains so; i.e., the end of the freeze-up period, and \textit{ice-off} as the first day on which a significant amount of water re-appears and remains visible for another day; i.e., the start of the break-up period.
\par
All the pixels that fall completely inside the lake boundary (obtained from \textit{openstreetmap.org}, generalised) are considered as \textit{clean} pixels and used for training and inference, so as to sidestep the handling of mixed pixels and the influence of geo-location errors.
Moreover, we call \textit{non-transition} days all (\textgreater30\% cloud-free) days on which the lake is either entirely frozen or entirely non-frozen. The remaining (again, \textgreater30\% cloud-free) days are referred to as \textit{transition} dates. Only non-transition days are used to train the segmentation, as spatially explicit labels for the transition days are difficult to obtain. Whereas the regression of the water fraction is trained on all the dates.
\section{Target lakes, winters, and satellites}\label{data_sec}
\subsection{Target lakes and winters}
\begin{figure*}[th!]
  \centering
  \includegraphics[width=0.99\linewidth]{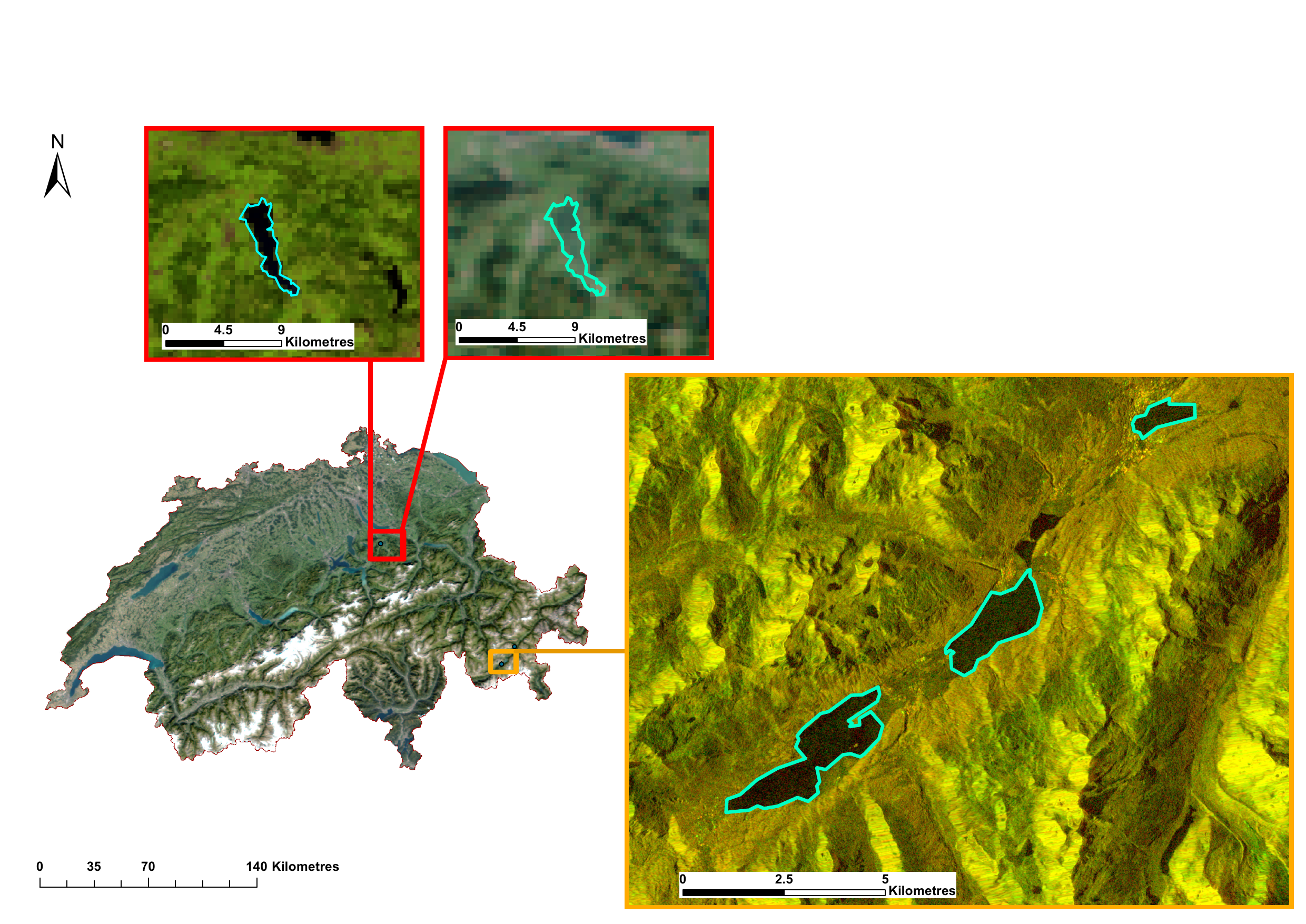}
  \caption{On the orthophoto map of Switzerland, the geographic locations of the target Swiss lakes are marked as red and amber rectangles. RGB True Colour Composite (TCC) of VIIRS (top left, R=$I_{3}$, G=$I_{2}$, B=$I_{1}$, captured on 7 September 2016) and MODIS (top right, R=$B_{1}$, G=$B_{4}$, B=$B_{3}$, captured on 7 September 2016) for the region Einsiedeln (lake Sihl) are shown in the zoomed red rectangles. For the region Engadin (lakes Sils, Silvaplana and St.~Moritz, from left to right), the RGB TCC (R=VV, G=VH, B=0) of Sentinel-1 SAR (captured on 13 September 2016) is shown in the zoomed amber rectangle. Best if viewed on screen.}
  \label{fig:geo_lakes}
\end{figure*}

For our case study, we monitor lake ice on four (mid-latitude, Alpine) lakes in Switzerland: Sihl, Sils, Silvaplana and St.~Moritz. See Fig.~\ref{fig:geo_lakes}. Lake Sihl is located near the Swiss plateau while the latter three lakes are located close to each other at higher altitudes in the Engadin valley, and share similar geographical and environmental conditions. More details about these lakes can be found in Table~\ref{table:lakes} (see Appendix~\ref{app:A}). For each lake, we process all three satellite sources for two winters: 2016--17 and 2017--18. In each winter, all available cloud-free dates between the beginning of September and the end of May are analysed.
\subsection{Target satellites}
\subsubsection{Optical satellite data (MODIS and VIIRS)}
We use Terra MODIS~\cite{TerraMODIS} and Suomi NPP VIIRS~\cite{SuomiNPP} data in our analysis. For both these data types, we follow the download and pre-processing procedures as described in Tom et al.~\cite{tom2020_RS_journal}, which include absolute geolocation correction~\cite{Sultan_2013}, back-projection of the generalised lake outlines onto the satellite images to extract clean pixels, cloud filtering, and bilinear interpolation for low-resolution bands (upsampling the 500m and 1km bands to 250m, only for MODIS). All cloud-free lake pixels in acquisitions with \textgreater30\% non-cloudy pixels are extracted and processed. We note that for VIIRS, after absolute geolocation correction, there is no clean pixel available for lake St.~Moritz.%
\footnote{We include this tiny lake mainly to assess the limits of satellite-based ice monitoring for very small lakes.} %
Instead of using all the available bands, 12~(5) out of 36~(22) spectral bands are selected for MODIS~(VIIRS) as suggested in Tom et al.~\cite{Tom_isprs_2018_optical}. For MODIS, this selection was done after a visual check and 12 potentially useful bands without artefacts (stripes) or saturation issues were picked. For VIIRS, all five imagery bands were selected. To generate images with uniform size (to be fed as input to the network), the lake pixels are padded with background pixels to form $12\!\times\!12$ patches (input dimensions of $12\!\times\!12\!\times\!12$, respectively $12\!\times\!12\!\times\!5$ for MODIS and VIIRS). Details about the optical images are displayed in Table~\ref{table:satellite_data}.
\subsubsection{Radar satellite data (Sentinel-1 SAR)}
We download already pre-processed dual polarisation C-band S1-SAR data as level-1 Ground Range Detected~(GRD) scenes in Interferometric Wide~(IW) swath mode from the Google Earth Engine~(GEE) platform~\cite{GEE_RSE_paper_2017}. The GEE pre-processing includes border and thermal noise removal, radiometric calibration, terrain correction, absolute geolocation correction, and log scaling. Additionally, we resize the images spatially to $128\!\times\!128$ resolution (covering all lake pixels and some background). Although the instrument can record four polarisations (VV, VH, HV, HH), only VV and VH are available for the regions of interest around Einsiedeln and Engadin, in Switzerland. Hence, the final input size for SAR images is $128\!\times\!128\!\times\!2$. The temporal resolution varies across regions as well as winters, ranging from 1.5-4.5 days. From winter 2017--18, the data is available every 1.5 days for lakes in region Engadin (Sils, Silvaplana, St.~Moritz) and every 3 days for region Einsiedeln (Sihl). On the other hand, in 2016--17, it is 2.3 and 4.5 days respectively. Sentinel-1 (S1) scans region Engadin in four orbits, but region Einsiedeln only in two orbits. This explains why the temporal resolution for Engadin is relatively high for a given winter. In winter 2017--18, the data from both S1A and S1B satellites are available. Additionally, even though the S1B was launched in April 2016, the corresponding data is available only since March 2017, effectively offering a relatively lower temporal resolution for winter 2016--17. More details on the S1-SAR data that we use are also summarised in Table~\ref{table:satellite_data}.
\subsubsection{Satellite combination - effective temporal resolution}
For the target lakes and winters, we compute the ETR (average number of days between cloud-free image acquisitions) separately for each input sensor and the combination, see Table \ref{table:temp_res}. The ETR value for the sensor combination indicates the limit to which temporal resolution can be improved by fusing the three sensors. On the other hand, the ETR value for each optical sensor shows the realistic temporal resolution achievable in the presence of clouds. For the optical satellites, only the cloud-free (at least 30\%) days are counted while computing the ETR. 
In any winter, there can be days on which all three satellites have imaged the regions of interest, days on which these regions were scanned more than once by the same satellite (rare), and days with no S1-SAR acquisition and no available cloud-free MODIS and VIIRS data. 
\begin{table}[th!]
\centering
\begin{tabular}{ccccccccc}
    \hline
    \textbf{Lake} & \textbf{Winter} & \textbf{S} & \textbf{M} & \textbf{V} & \textbf{S+M+V} \\
    \hline
    \multirow{2}{*}{Sihl} & 2016--17 & 4.5 & 1.9 & 2.1 & 1.5 \\  
                                & 2017--18 & 3.0 & 2.2 & 2.4 & 1.5 \\ \hline
    \multirow{2}{*}{Sils} & 2016--17 & 2.3 & 1.7 & 1.7 & 1.3\\ 
                                & 2017--18 & 1.5 & 2.0 & 2.0 & 1.2 \\ \hline
    \multirow{2}{*}{Silvaplana} & 2016--17 & 2.3 & 1.7 & 1.7 & 1.3 \\ 
                                & 2017--18 & 1.5 & 2.1 & 1.9 & 1.3  \\ \hline
    \multirow{2}{*}{St.~Moritz} & 2016--17 & 2.3 & 1.7 & - & 1.4 \\ 
                                & 2017--18 & 1.5 & 2.0 & - & 1.3 \\ \hline
  \end{tabular}
  \caption{Effective Temporal Resolution (ETR) during the target winters in days, for different sensors. S, M, and V denote S1-SAR, MODIS and VIIRS respectively.}
  \label{table:temp_res}
 \end{table} 
\par
For a given lake and winter, different ETR for MODIS and VIIRS is possible since their overpasses occur at different times of day and the cloud patterns may change. Additionally, in the presence of scattered clouds, the cloud cover can vary even between the lakes located in the same valley (Sils, Silvaplana, St.~Moritz), in the same overpass. Even though both optical satellites have daily acquisitions, the ETR for each sensor is greater than 1 day even at a low cloud threshold (at least 30\% non-cloudy pixels). In the best case with zero clouds, both MODIS and VIIRS individually would have daily temporal resolution and no fusion with SAR data is needed. However, in practice, clouds are inevitable in the target regions and the chosen scenario of using (S+M+V) is a realistic minimal setup:
For the two winters of interest, the ETR after combining the data from three sensors is between 1 and 2 days. See Table~\ref{table:temp_res} (a day is counted only once if captured by more than one satellite).  
It can be noted from Table \ref{table:temp_res} that winter 2017--18 was relatively more cloudy than 2016--17. Still, the temporal resolution of the three sensors combined is better in 2017--18, due to the extra data from the S1B satellite, which was not available in 2016--17. 
\subsection{Ground truth}\label{sec:GT}
The reference data was generated by visual interpretation of the images from freely available webcams located in the vicinity of the target lakes. The images were labelled independently by a human operator and independently confirmed by a second one. Whenever there were doubts about the per-day label, the operators used more images from the same webcam, other webcams monitoring the same lake (if available), information from adjacent days, and in some cases also the online media reports and interpretation of S2 satellite images (if available and cloud-free) to corroborate their final choice of label. On non-transition days, only a binary label was assigned: fully-frozen (fraction of water pixels 0) or fully non-frozen (fraction 1). Whereas on transition days the assigned labels have higher granularity and also include the states: more frozen (0.25) and more non-frozen (0.75). There is a price to pay, as these additional states cannot always be labelled reliably and a significant amount of label noise must be expected. Due to interpretation uncertainties and not clearly visible regions, as well as the fact that not the entire lake surface is visible in the webcam images, the actual water area of a "more frozen" lake is likely to fluctuate in the range 60-90\% of the lake surface. Moreover, due to interpretation errors caused by bad lighting, compression artefacts of webcam streams, and low spatial resolution especially in the far-field, occasional confusions between adjacent states are almost certainly present in the reference data. These issues can be hardly avoided, as webcams are typically placed relatively low onshore, such that the lake surface is viewed at acute angles and the image scale degrades rapidly with distance from the camera.

For training the segmentation task, only the non-transition dates are used, after converting the per-image labels to spatially explicit maps with 100\% frozen, respectively non-frozen pixels inside the lake boundary. Pixels outside the lake are labelled as background. 
\section{Methodology}\label{method}
\subsection{Data fusion model}
\begin{figure*}[!]
    \centering
    \includegraphics[width=0.99\linewidth]{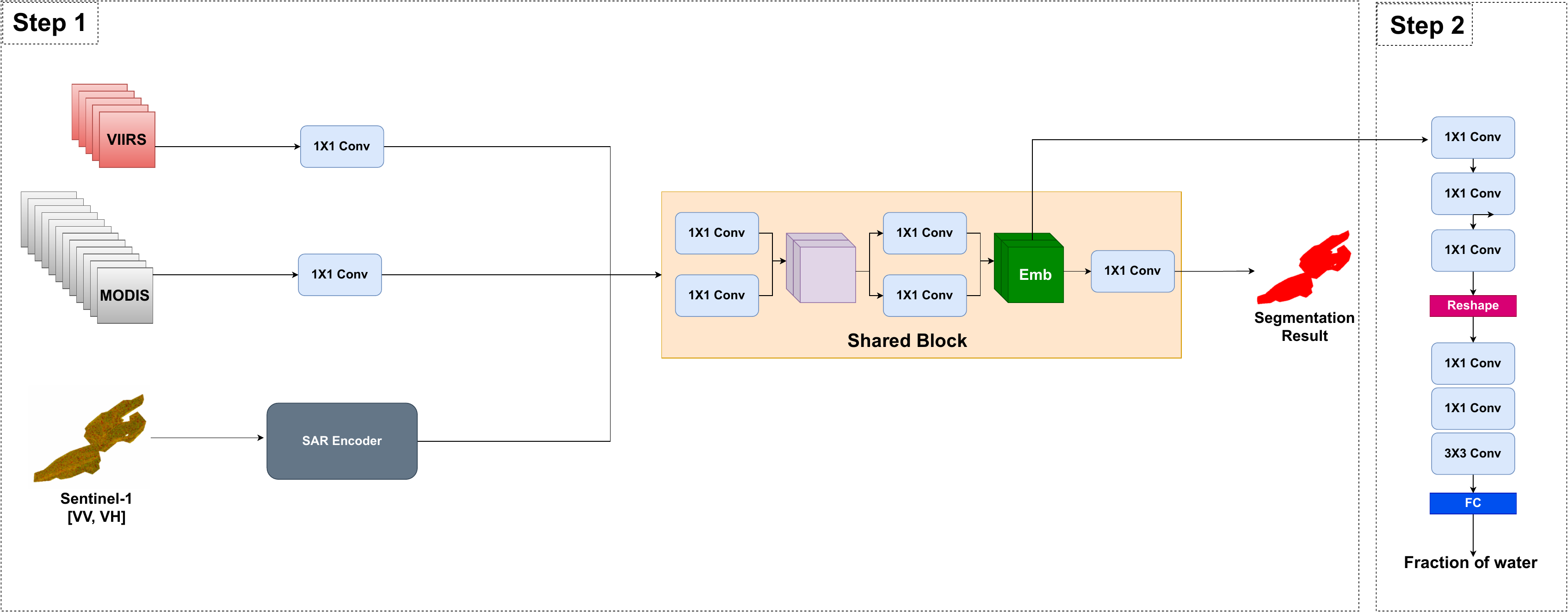}
    \caption{Network structure of the proposed \textit{2-step} model. In step 1, the shared block generates the joint embedding space (Emb). In step 2, the embedding learnt are fused to learn the per-day fraction of non-frozen (water) pixels. Effectively, two outcomes are predicted by the proposed model: 1) in step 1, the per-pixel semantic segmentation of the lake (red=\textit{non-frozen}, blue=\textit{frozen}, here we have shown an example non-frozen result for lake Sils), 2) in step 2, the point estimate of the daily fraction of water pixels. \textit{Conv} and \textit{FC} represent the convolution and fully connected layers respectively. Best if viewed on screen.}
    \label{fig:2step}
\end{figure*}

We propose a \textit{2-step} model for deep satellite data fusion, see Fig.~\ref{fig:2step} for a depiction of the network architecture. The two underlying steps are:
\begin{enumerate}
    \item learn a joint representation (feature space) that can embed data from multiple satellite sensors
    \item use the learnt embedding as input for the interpretation task, in our case lake ice monitoring
\end{enumerate}
The first step transforms the inputs into a new feature space (a "latent representation"). To train that step, which should preserve and accentuate information about the local state of the lake surface, we use the auxiliary task to explicitly segment the lake into three semantic classes, \textit{frozen}, \textit{non-frozen} and \textit{background}. The second step starts from the resulting feature representation and regresses the fraction of water (non-frozen) pixels on the lake, including a mechanism for multi-temporal analysis over a short period around the day in question.

In the following, we describe each of the two steps in more detail.
\vspace{1em}
\subsubsection*{\bf Step 1: Joint embedding}\label{sec:2-step_step1}%
To perform fusion, the domain gap between the heterogeneous inputs that stem from satellite sensors operating at very different wavelengths (optical, radar) and spatial resolutions (20 to 375m), need to be bridged. We achieve this by learning a shared, intermediate representation, along with individual encoders (a.k.a.\ "embedding function") to transform the data of each sensor to that representation (the common "embedding", abbreviated as "Emb" in Fig.~\ref{fig:2step}).
The overall architecture consists of an individual branch per sensor to map different inputs to a shared "feature map" and a shared block with the same weights for all sensors to further abstract that feature map into the final representation that serves as the basis for the output predictions.
\par
\begin{figure*}[!]
    \centering
    \includegraphics[width=0.75\linewidth]{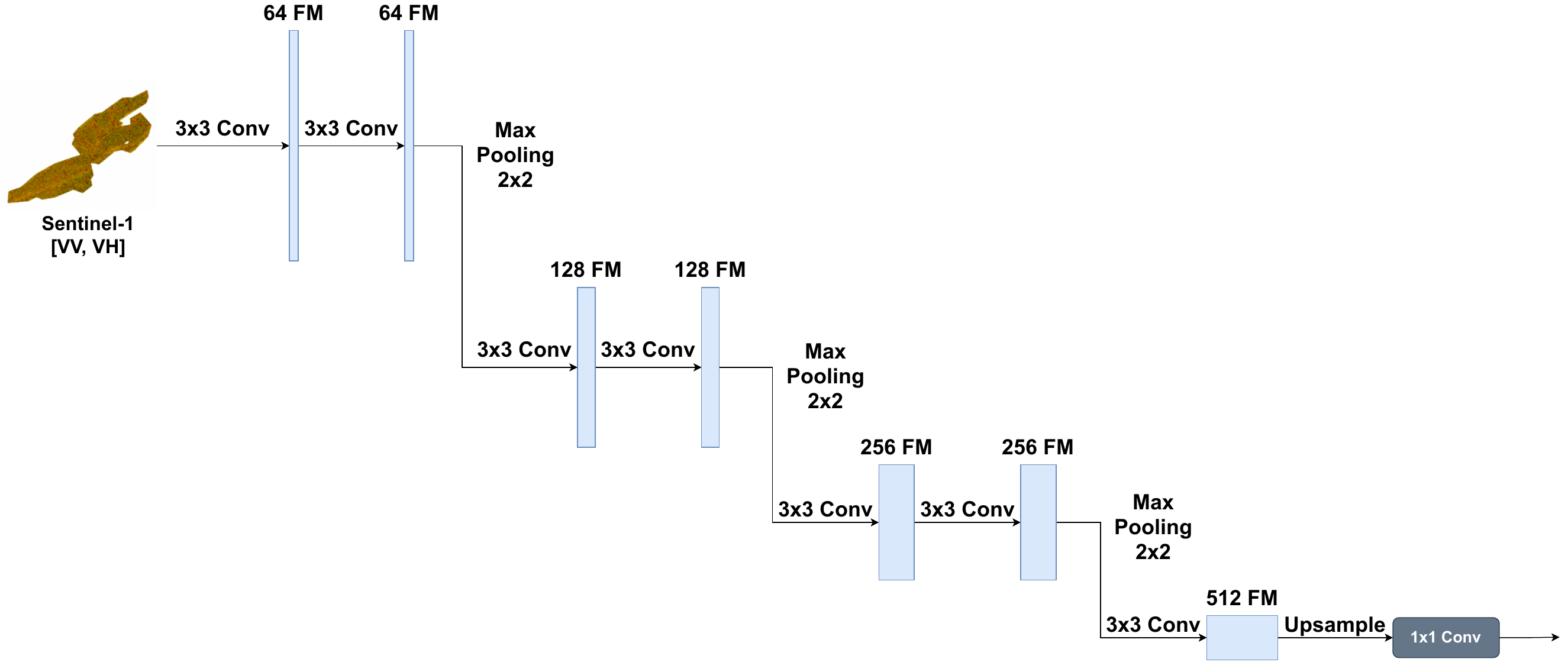}
    \caption{U-net style SAR encoder architecture. Conv and FM represent convolution layer and feature maps respectively. Best if viewed on screen.}
    \label{fig:sar_encoder}
\end{figure*}

\textit{Encoders: }
At first, the input image is encoded into intermediate features that, on the one hand, can be derived from any of the input sensors and, on the other hand, preserve the information needed to differentiate the frozen and non-frozen states. For the MODIS and VIIRS branches, with their low spatial resolutions that do not call for texture analysis, we use a simple $1 \times 1$ convolution (Conv) layer as the encoder. On the contrary, the S1-SAR branch has a much higher spatial resolution and must be able to learn texture and context features over spatial neighbourhoods of multiple pixels. Our encoder is a CNN inspired by U-net~\cite{RonnebergerFB15}, with leaky ReLU~\cite{leakyReLU2015} activation. See Fig.~\ref{fig:sar_encoder}.

\par
\textit{Shared block: }The per-sensor branches, which one might interpret as a "pre-processing" into a common feature space, are followed by a shared block to optimise the features for the downstream task of lake ice monitoring. That block consists of a shallow network with $1\!\times\!1$ convolution and concatenation layers, see Fig.~\ref{fig:2step}. Also, in this block, all weights are learnt from data. Its output is the deep embedding (Emb), with a uniform size of $12\!\times\!12\!\times\!32$. It serves as input for the second step, described in Section~\ref{sec:2-step_step2}. 
\par 
\textit{Training: }The network weights in the first step of the 2-step model are trained on the auxiliary task of spatially explicit semantic segmentation, as follows:
\begin{itemize}
    \item First, the weights in the U-net style SAR encoder are pre-trained, starting from "Glorot" uniform initialisation~\cite{Glorot2010}, by feeding the network with S1-SAR data. The leakage rate (slope for values \textless0) of the leaky ReLU is set as $\alpha=0.1$.
    \item Second, the MODIS and VIIRS encoders are pre-trained, along with the shared block, again with "Glorot" uniform initialisation~\cite{Glorot2010}. Optical images are fed in an alternating fashion, switching between one epoch with MODIS data and one epoch with VIIRS data. This procedure is run for 40 epochs (20 each for MODIS and VIIRS), thus priming the two branches and the shared block for MODIS-VIIRS fusion.
    \item Third, the shared block weights are fine-tuned to make them compatible also with SAR data. To that end the network is again fed with S1-SAR data, thus jointly updating the pre-trained SAR encoder and shared block. This last step tunes the shared block to also perform optical-SAR (MODIS-SAR, VIIRS-SAR) fusion.
\end{itemize}
\par
In all above-mentioned training steps, the weights are optimised by minimising the \textit{cross entropy loss} over pixel-wise class labels, using the \textit{Adam} optimiser~\cite{kingma2014adam}. Details about hyper-parameters of the training are given in Table~\ref{table:2step_nw_parameters}, columns 2-4.
\par
In the proposed setup, the type and number of input branches are flexible. It is fairly straightforward to include further satellite sensors, by designing a suitable encoder and fine-tuning the shared block to accommodate the new input data characteristics. In that context note that the convolution layer can handle varying input sizes, so input images of varying sizes, respectively spatial resolution, can be used (in fact our inputs also do not have the same pixel dimensions).
\begin{table*}[th]
\centering
\resizebox{0.9\textwidth}{!}{%
\begin{tabular}{c|ccc|c}
\toprule
 &  & \multicolumn{3}{c}{\textbf{2-step model}}\\
 & \textbf{step 1} & \textbf{step 1} & \textbf{step 1} & \textbf{step 2} \\
 & uNet-style SAR & MODIS, VIIRS encoder pre-training, & Shared block & Regression \\
 & encoder pre-training & Shared block pre-training & fine-tuning & \\
\midrule
\textit{Epochs} & 500 & MODIS (20), VIIRS (20) & 250 & 100\\
\textit{Batch size} & 16 & 8 & 16 & 4 \\
\textit{Loss} & cross entropy & cross entropy & cross entropy & $L_{reg}$ \\
\textit{Optimiser} & Adam~\cite{kingma2014adam} & Adam~\cite{kingma2014adam} & Adam~\cite{kingma2014adam} & SGD \\
\textit{LR} & 5e-05 & 5e-04 &  1e-05 & 5e-04 \\
\textit{LR decay} & 375, 0.9 & - & 150, 0.9 & -\\
\textit{(steps, rate)} &  &  &  &  \\
\textit{window size (w)} & - & - & - & 7 \\
\bottomrule
\end{tabular}}
\caption{Network parameter settings for various training stages in the proposed \textit{2-step} model. Learning Rate (LR) decay refers to exponential scheduling. SGD represents stochastic gradient descent.}
\label{table:2step_nw_parameters}
\end{table*} 
%
%
\vspace{1em}
\subsubsection*{\bf Step 2: prediction based on the learnt embedding}\label{sec:2-step_step2}
The second step in our application task is a regression model that predicts what fraction of the lake surface is covered by open water. This prediction is based on embedding from multiple adjacent days, to exploit temporal sequence information and to smooth out noise in the per-day features. The regression network starts with three convolutional layers that operate independently on each per-day feature map, followed by a reshape layer that aligns the information from a \emph{window} of adjacent time steps (i.e., the time and channel dimensions are collapsed into a single dimension). Three more convolution layers then combine the information from all time steps in the window, and a final, fully connected (FC) layer returns a single, scalar prediction, assigned to the central day in the window. See Fig.~\ref{fig:2step},
\par
The \textit{window size} is a hyper-parameter that determines how many per-day embedding are stacked into a local time-series. E.g., when it is set to 7, a daily prediction is based on feature maps from the previous 3, current and next 3 days.
\par
For the regression model, the loss function $L_{reg}$ is defined as
\begin{equation}
    L_{reg} = L_{mse} + \beta L_{line} + \gamma L_{idc}
    \label{L_reg_equation}
\end{equation}
where $L_{mse}$, $L_{line}$ and $L_{idc}$ correspond to \textit{mean squared error loss}, \textit{line loss}, and \textit{intra-day coherence loss} respectively. We empirically fixed the weights $\beta=0.25$ and $\gamma=0.08$. In step 2 also, the training starts from "Glorot" uniform weight initialisation~\cite{Glorot2010}. Further training settings for the regression model is shown in Table~\ref{table:2step_nw_parameters}~(column 5).
\par
$L_{line}$ penalises the deviations from a linear trend ($mx+c$), computed locally for each mini-batch of size $b$. Each batch is carefully selected so that the embedding from adjacent days are present. The line loss is based on the assumption that the predictions of adjacent dates should lie on a straight line, which acts as a smoothness prior and mainly helps to correct isolated outliers. First, the line is fitted as
\begin{equation}
    m = \frac{y^{l} - y^{f}}{b},
    \; \; 
    c = y^{f}
\end{equation}
with $y^{l}$, $y^{f}$ the last and first prediction in the mini-batch. The deviation $d^i$ of the $i^\text{th}$ prediction $y^i$ from the line is then
\begin{equation}
    d^i = \frac{| mi+c-y^i|}{\sqrt{m^2+1}}
\end{equation}
\par
There are a few days when more than one satellite imaged the region of interest. In such cases, multiple embedding (one per satellite) will be generated by step 1. The loss term $L_{idc}$ computes the variance between predictions from the same day and penalises it, thereby encouraging coherent outputs. The final output (fraction of water) for a day with more than one sensor observation is the mean of the individual predictions.
\section{Experiments, results and discussion}\label{experiment}
We perform experiments in the leave-one-out setting: the dataset is sub-divided into different portions, and in each run, one portion is left out and a model is trained on the remaining data. The model thus trained is then tested using the left out portion. We use two settings: \textit{Leave One Winter Out} (LOWO) and \textit{Leave One Lake Out} (LOLO), to assess the model's capability to generalise across time and space. For instance, "LOLO-Sihl" means training on the data of all lakes except Sihl (from both winters 2016--17 and 2017--18) and testing on the data of Sihl from both winters, and similar for other combinations. 
\par
Using the proposed \textit{2-step} model, we generate embedding on all available dates, from both non-transition and transition periods. To train step 1 (semantic segmentation) which requires per-pixel labels, we use only data from non-transition days, because pixel-wise ground truth is not available for the transition days. Step 2 (regression of water fraction) needs only scalar per-day labels, hence we include data from all days in the training set, despite the inevitably higher label noise during transitions (see Section~\ref{sec:GT}).
\par
Although an auxiliary task in our setup, we quantitatively assess segmentation performance, in terms of overall classification accuracy and mean per-class Intersection-over-Union (mIoU) score, see Section \ref{sec:quant_semseg}. For qualitative assessment, we visualise the learnt embedding in Section~\ref{sec:qualitative}. As an additional check we plot and visually inspect the time-series of predicted water fractions for each lake throughout the entire winter, see Section~\ref{sec:qualitative:fw}. From those time-series, we also derive the ice-on/-off dates and assess the model's ability to retrieve lake ice phenology, in Section~\ref{sec:ice-onoff}. 
\subsection{Quantitative results: semantic segmentation}
\label{sec:quant_semseg}
To learn the embedding, our model is trained to perform pixel-wise semantic segmentation, which we assume to be a good proxy task for the retrieval of ice cover, but with a stronger, spatially more explicit and more detailed supervision signal. This experiment inspects the performance of the segmentation step (step 1). After training the model, the data from each satellite sensor is fed to the respective input branch and is analysed independently, for two reasons. On the one hand, it allows us to separately quantify the performance of the model for each satellite sensor. On the other hand, the embedding is trained jointly, but still computed features separately for each input image, and thus also for each sensor -- the mixing of features extracted from different sources only happens when aggregating over time windows in the regression step. LOWO and LOLO results are shown in Tables \ref{tab:segment_LOWO} and \ref{tab:segment_LOLO}, respectively. To quantify the epistemic uncertainty (model uncertainty due to imperfect training data), we also estimate the standard deviation of the predictions, using an ensemble of five models with independent random initialisations. The results support our assumption that a joint model with a single, shared embedding can handle inputs from any single sensor. In more detail, it can be seen that the proposed model achieves very good generalisation across different winters (mIoU \textgreater76.1\%, accuracy\textgreater94.6\%). Overall, also the generalisation across different lakes is very good for optical images (mIoU \textgreater79.8\%, accuracy\textgreater91.6\%), with a small decrease for the S1-SAR data (mIoU \textgreater60.7\%, accuracy\textgreater91.3\%).
\begin{table*}[!]
\centering
\small
\begin{tabular}{ccccccccc}
\toprule
\textbf{Data} &  \multicolumn{4}{c}{\textbf{Winter 2016--17}} & \multicolumn{4}{c}{\textbf{Winter 2017--18}} \\
{}    & \multicolumn{2}{c}{\textit{Accuracy}} & \multicolumn{2}{c}{\textit{mIoU}} & \multicolumn{2}{c}{\textit{Accuracy}} & \multicolumn{2}{c}{\textit{mIoU}} \\
\textit{} & $\mu$ & $\sigma$ & $\mu$ & $\sigma$ & $\mu$ & $\sigma$ & $\mu$ & $\sigma$\\
\midrule
\textit{MODIS} &  96.1 & \textcolor{gray}{0.1} & 81.2 & \textcolor{gray}{0.4} & 97.5 & \textcolor{gray}{0.5} & 84.9 & \textcolor{gray}{1.7} \\
\textit{VIIRS} & 98.9 & \textcolor{gray}{0.2} & 87.9 & \textcolor{gray}{1.9} & 99.1 & \textcolor{gray}{0.1} & 87.4 & \textcolor{gray}{1.1} \\
\textit{S1-SAR} & 94.6 & \textcolor{gray}{0.3} & 76.1 & \textcolor{gray}{1.0} & 95.2 & \textcolor{gray}{0.1} & 79.1 & \textcolor{gray}{0.4} \\
\bottomrule
\end{tabular}
\caption{Semantic segmentation results (in \%) of \textit{leave one winter out} experiment. $\mu$ and $\sigma$ denote the means and empirical standard deviations across five models trained with different random initialisations.}
\label{tab:segment_LOWO}
\end{table*} 
\begin{table*}[!]
\centering
\small
\begin{tabular}{ccccccccccccccccc}
\toprule
\textbf{Data} & \multicolumn{4}{c}{\textbf{Sihl}} & \multicolumn{4}{c}{\textbf{Sils}} &  \multicolumn{4}{c}{\textbf{Silvaplana}} 
&  \multicolumn{4}{c}{\textbf{St.~Moritz}} \\
{}   & \multicolumn{2}{c}{\textit{Accuracy}} & \multicolumn{2}{c}{\textit{mIoU}} & \multicolumn{2}{c}{\textit{Accuracy}} & \multicolumn{2}{c}{\textit{mIoU}}
& \multicolumn{2}{c}{\textit{Accuracy}} & \multicolumn{2}{c}{\textit{mIoU}} & \multicolumn{2}{c}{\textit{Accuracy}} & \multicolumn{2}{c}{\textit{mIoU}}\\
\textit{} & $\mu$ & $\sigma$ & $\mu$ & $\sigma$ & $\mu$ & $\sigma$ & $\mu$ & $\sigma$ & $\mu$ & $\sigma$ & $\mu$ & $\sigma$ & $\mu$ & $\sigma$ & $\mu$ & $\sigma$\\
\midrule
\textit{MODIS} & 91.6 & \textcolor{gray}{2.5} & 79.8 & \textcolor{gray}{3.3} & 98 & \textcolor{gray}{0.2} & 89.2 & \textcolor{gray}{0.9} & 98.1 & \textcolor{gray}{0.3} & 85.7 & \textcolor{gray}{2.0} & 99.8 & \textcolor{gray}{0} & 80.9 & \textcolor{gray}{1.6}\\
\textit{VIIRS} & 98.3 & \textcolor{gray}{0.3} & 87.8 & \textcolor{gray}{1.8} & 99.5 & \textcolor{gray}{0.1} & 90.9 & \textcolor{gray}{1.3} & 99.5 & \textcolor{gray}{0.1} & 89.3 & \textcolor{gray}{1.6} & - & - & - & - \\
\textit{S1-SAR} & 91.3 & \textcolor{gray}{0.6} & 60.7 & \textcolor{gray}{1.4} &	92.2 & \textcolor{gray}{0.5} & 70.3 & \textcolor{gray}{1.4} & 92.5 & \textcolor{gray}{0.4} & 72.6 & \textcolor{gray}{1.1} & 93.9 & \textcolor{gray}{0.3} & 72.8 & \textcolor{gray}{1.1} \\
\bottomrule
\end{tabular}
\caption{Semantic segmentation results (in \%) of \textit{leave one lake out} experiment. $\mu$ and $\sigma$ denote the means and empirical standard deviations across five models trained with different random initialisations.}
\label{tab:segment_LOLO}
\end{table*} 
\subsection{Qualitative results: full winter time-series}\label{sec:qualitative:fw}
\begin{figure*}[!]
    \centering
    \subfloat[\centering Lake Sihl]{{\includegraphics[width=.99\textwidth]{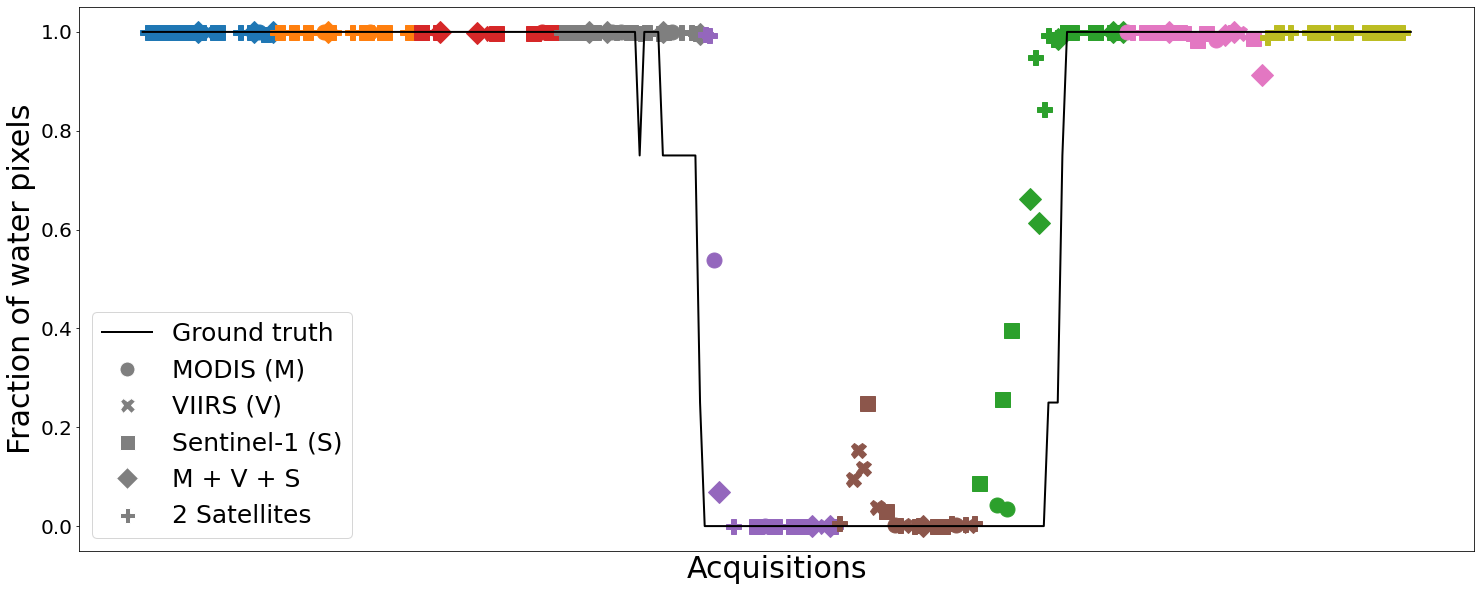}}}\\
    \subfloat[\centering Lake Sils]{{\includegraphics[width=.99\textwidth]{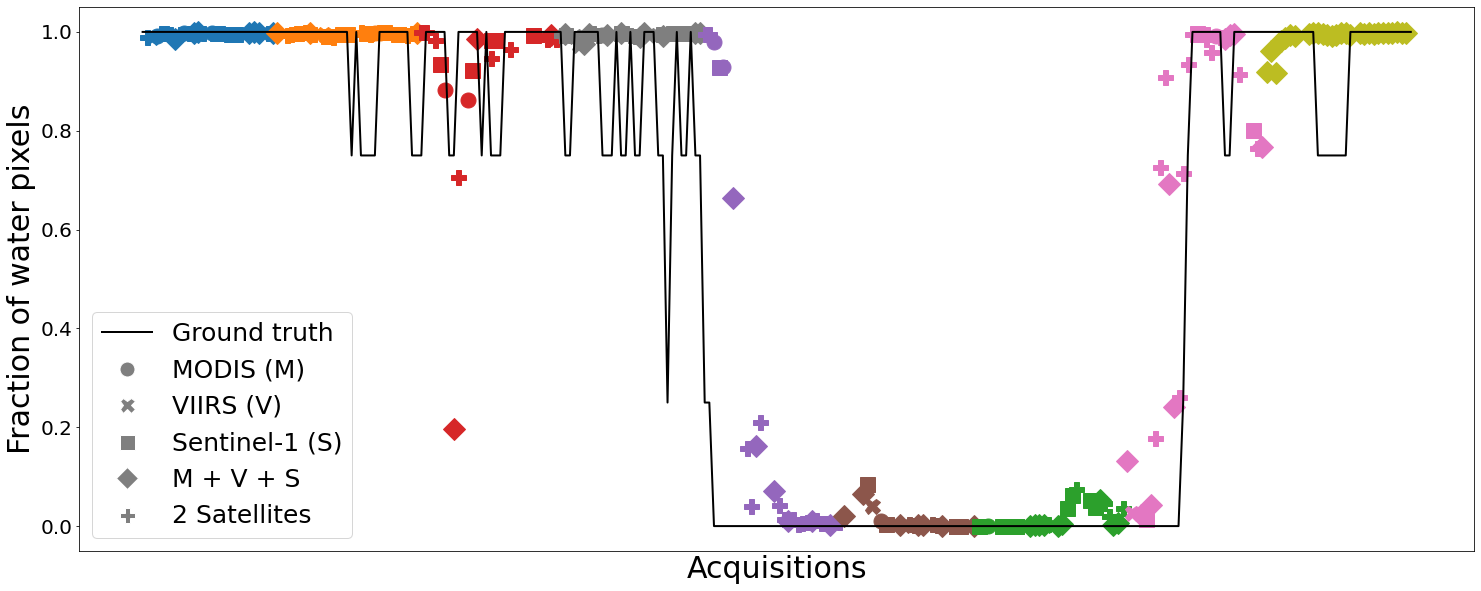}}}
    \caption{Time-series plots for lakes Sihl and Sils from winter 2016--17 using a model trained on all the data (all four lakes) from winter 2017--18. Predictions are separated by colors month-wise. Best if viewed on screen.}%
    \label{fig:time_seris_main_sihl}%
\end{figure*}

In each winter, after fusing the embedding from all three satellite sources, daily predictions of the fraction of water (non-frozen pixels) are obtained for each lake and the time-series is generated. Fig.~\ref{fig:time_seris_main_sihl} shows the sample results for lakes Sihl and Sils. Here, the model is trained using the data from all the lakes from winter 2017--18. In each sub-figure, the predicted fraction of water is shown on the $y$-axis against the acquisitions (dates) on the $x$-axis displayed in chronological order from the beginning of September until the end of May. Reference annotations are shown as a black line, and predictions are displayed as points, with a different colour per calendar month. Different markers denote the three different satellite sensors, days on which a lake was captured by all three sensors (M+V+S), and (M+V, M+S or V+S, denoted as "2 satellites" and sharing the same marker). Results for the remaining two lakes are shown in Appendix~\ref{appendix:B}. 
\par
It can be inferred from Figs.~\ref{fig:time_seris_main_sihl} and \ref{fig:time_series_2step_silvaplana_stmoritz} that the time-series results are good for the bigger lakes Sihl and Sils, satisfactory for Silvaplana, but not very good for lake St.~Moritz, especially during freeze-up.
As said earlier, that lake is tiny (not a single clean pixel in VIIRS, four clean pixels in MODIS) and has been included as an extreme case to test the limits of low spatial-resolution satellite data for small lakes.
As expected, the results generally improve with increasing lake size, see also Table~\ref{table:lakes} (Appendix~\ref{app:A}) and Figs.~\ref{fig:ablation:aux_loss} and \ref{fig:ablation:win_size}. There are several possible reasons. For low-resolution sensors, there are simply few pixels to learn the model, whereas for higher-resolution sensors like S1-SAR there may be enough pixels, but still only little context. Moreover, the portion covered by mixed pixels along the lake boundary is relatively larger for low spatial resolution sensors, and such boundary effects are compounded by the fact that the mixed pixels are most affected by residual geo-localisation errors. Moreover, quantisation and correspondence issues may play a role when transferring ground truth annotations from webcams to satellite images, see also Section \ref{sec:GT}. Finally, note that for St.~Moritz only two sensors were fused as there are no clean VIIRS pixels.
\subsection{Qualitative visualisation: learnt embedding}\label{sec:qualitative}
\begin{figure*}[!]
    \centering
    \subfloat[Lake Sihl,  dimension=3]{\includegraphics[width=.49\linewidth]{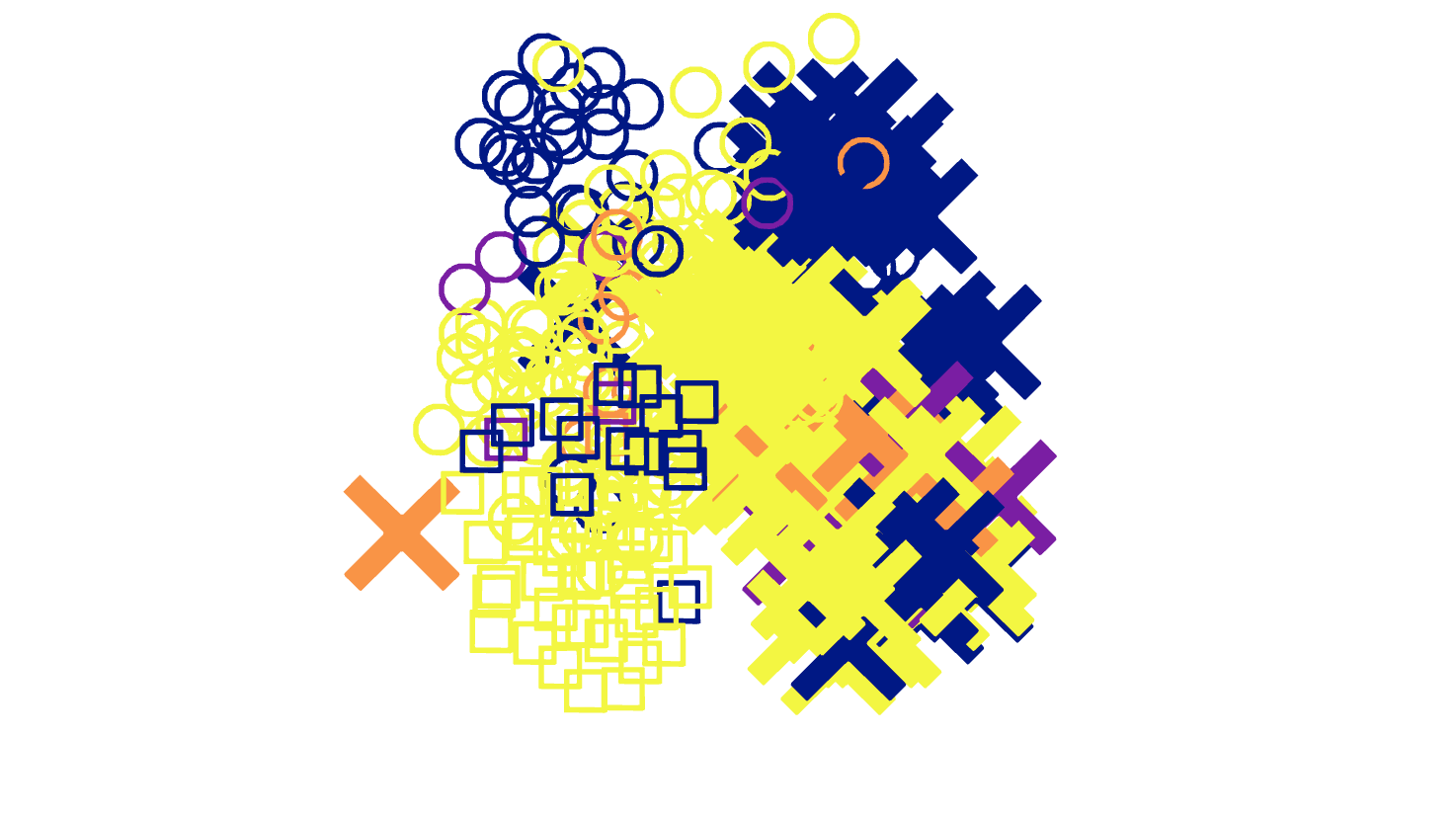}}
    \subfloat[Lake Sihl,  dimension=2]{\includegraphics[width=.49\linewidth]{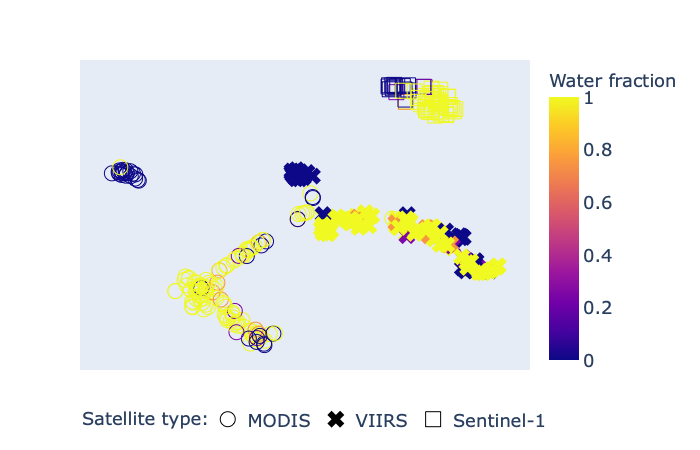}}\\
    \subfloat[Lake Sils,  dimension=3]{\includegraphics[width=.49\linewidth]{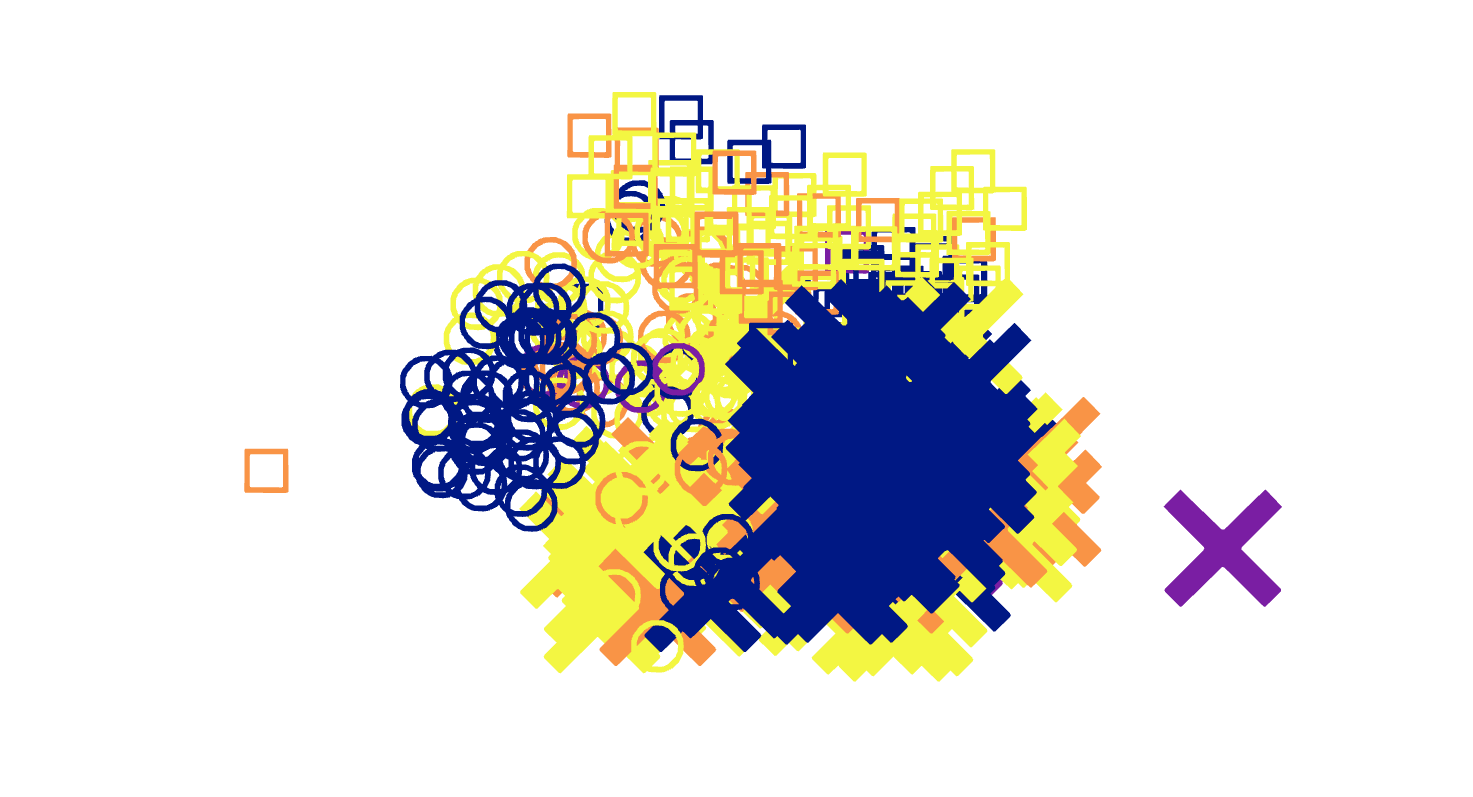}}
    \subfloat[Lake Sils,  dimension=2]{\includegraphics[width=.49\linewidth]{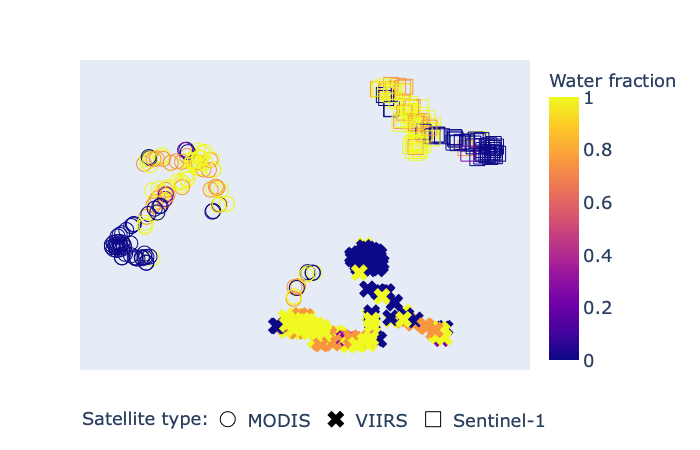}}\\
    \caption{$t$-SNE representation of the embedding learnt (lakes Sihl and Sils, winter 2016--17) using the proposed \textit{2-step} approach with a model trained on all the data from winter 2017--18. Water fraction refers to the fraction of non-frozen pixels. Different marker types are used to distinguish embeddings of different sensors. Best if viewed on screen.}%
    \label{fig:emb_vis}%
\end{figure*}

We visualise the learnt embedding after dimension reduction with $t$-distributed Stochastic Neighbour Embedding ($t$-SNE)~\cite{t-SNE}, for a visual impression of how the data cluster in high-dimensional feature space. See Fig.~\ref{fig:emb_vis}, where each dot corresponds to the feature embedding of a single acquisition (entire image from a single overpass of one sensor). In the illustration, blue corresponds to a high likelihood of ice/snow, while yellow means a high likelihood of water. It can be seen that the learnt embedding looks reasonable, in the sense that the classes are separable. Reducing the dimension of the projection to only 2 (right column) reveals that clusters for different sensors can still be restored. I.e., the joint feature space makes it possible to construct decision boundaries, respectively regression coefficients, that are invariant w.r.t.\ the input type, but the features themselves are not completely invariant.
\subsection{Ice-on/off results}\label{sec:ice-onoff}
To compare the performance of the proposed approach with previous works which analysed the target lakes%
~(\cite{tom_isprs_2020_sar,tom2020_RS_journal,LIP1_final_report_2019}), we extract the ice-on/off dates from the daily predictions of water fraction, see Table~\ref{table:ice_onoff}. For robustness, those daily predictions are again ensemble over five independently initialised networks. Furthermore, we perform a comparison with the in-situ temperature analysis reported in \cite{LIP1_final_report_2019}. The comparisons are only possible for winter 2016--17, as no other estimates are available for 2017--18.
\par
Following~\cite{Tom2021_RSE_submission}, we opt for a higher freeze threshold of 30\% to estimate the two LIP events. I.e., ice-on is detected when \textless30\% of the lake is not frozen, and ice-off when \textgreater30\% of it returns to the non-frozen state after the freeze period. The choice of threshold generally does not make a big difference, but the more conservative setting of 30\% is sometimes beneficial to compensate for the higher uncertainty in remote sensing estimates, as opposed to in-situ observations. For completeness, we also show the results for a threshold value of 10\%. See Table~\ref{table:ice_onoff}.
\par
In the Table, "M+V" refers to a decision-level fusion of MODIS and VIIRS predictions obtained with Support Vector Machines, as reported in~\cite{tom2020_RS_journal}. Similarly, "Webcam" means the prediction results from webcam images, obtained with the \textit{Deep-U-Lab} network~\cite{tom2020_RS_journal}. The ground truth is determined by interactive visual interpretation. Outputs that meet the GCOS target of $\pm$2 days (relative to the ground truth annotations) are printed in bold.
\begin{table*}[th]
\small
\centering
\begin{tabular}{ccccccccccc}
\toprule
\textbf{Date} & \textbf{Ground} & \textbf{\textit{2-step}} & \textbf{\textit{2-step}} &  \textbf{M+V} & \textbf{Webcam} & \textbf{S1-SAR} & \textbf{In-situ (T)}\\
\textbf{} & \textbf{truth} & \textbf{\textit{(10\%)}} & \textbf{\textit{(30\%)}} &  \textbf{\cite{tom2020_RS_journal}} & \textbf{\cite{tom2020_RS_journal}} & \textbf{\cite{tom_isprs_2020_sar}} & \textbf{\cite{LIP1_final_report_2019}}\\
\midrule
\textit{Ice-on (Sihl)}  & 1 Jan & \textbf{3 Jan} & \textbf{3 Jan} & \textbf{3 Jan} & 4 Jan & 11 Jan & 28-29 Dec\\
\textit{Ice-off (Sihl)} & 14 Mar & 6 Mar & 10 Mar & 10 Mar & 14 Feb & 28 Feb & \textbf{16 Mar} \\
\textit{} & 15 Mar &  &  &  &  & & \\
\textit{Ice-on (Sils)} & 2 Jan & 15 Jan & 9 Jan & \textbf{6 Jan} & - & \textbf{6 Jan} & \textbf{31 Dec} \\
\textit{} & 5 Jan &  & 15 Jan &  &  &  & \\
\textit{Ice-off (Sils)} & 8 Apr & 31 Mar & \textbf{7 Apr} & 31 Mar & - & \textbf{12 Apr} & \textbf{10 Apr} \\
\textit{} & 11 Apr & \textbf{6 Apr} & \textbf{12 Apr} &  &  \\
\textit{} &  & \textbf{11 Apr} &  &  &  &  & \\
\textit{Ice-on (Silvaplana)} & 12 Jan & 16 Jan & 15 Jan & 15 Jan & - & 15 Jan & \textbf{14 Jan} \\
\textit{Ice-off (Silvaplana)} & 11 Apr & 7 Apr & 7 Apr & 30 Mar & - & 18 Mar & 14 Apr \\
\textit{} &  & \textbf{12 Apr} & \textbf{13 Apr} &  &  & \\
\textit{Ice-on (St.~Moritz)} & 15-17 Dec & 17 Jan & 6 Jan & 1 Jan & \textbf{14 Dec} & 25 Dec & \textbf{17 Dec} \\
\textit{} & & 27 Jan & 16 Jan &  &  & \\
\textit{} &  &  & 25 Jan & &  & \\
\textit{Ice-off (St.~Moritz)} & 30 Mar - 6 Apr & \textbf{7 Apr} & \textbf{7 Apr} & \textbf{7 Apr} & 18 Mar-26 Apr & \textbf{30 Mar} & \textbf{5-8 Apr}\\
\bottomrule
\end{tabular}
\caption{Ice-on/off dates (winter 2016--17) estimated with the \textit{2-step} model, with two different thresholds. For comparison we show the ground truth (in chronological order when more than one candidate exists), the earlier remote sensing results~(\cite{tom_isprs_2020_sar,tom2020_RS_journal}), and the results of in-situ (temperature, T) analysis~\cite{LIP1_final_report_2019}. Results that meet the GCOS requirement are printed bold.}
\label{table:ice_onoff}
\end{table*} 
\par
When comparing the results in Table~\ref{table:ice_onoff} from different satellite remote sensing methods~(\cite{tom2020_RS_journal,tom_isprs_2020_sar}) methods, the joint embedding model, on average, deviates least from the ground truth. However, the results of in-situ temperature analysis~\cite{LIP1_final_report_2019}, are better. As before, the predictions for the two (relatively) larger lakes Sihl and Sils are more accurate. The accuracy is inversely proportional to the lake area, see also Table~\ref{table:lakes} (Appendix~\ref{app:A}). For the tiny lake St.~Moritz, all satellite-based predictions of ice-on are wildly off, again showing the limits of satellite remote sensing and/or statistical machine learning for very small geographic objects.
\subsection{Runtime} On a computer equipped with NVIDIA GeForce GTX 1080 Ti graphics card (12 GB), it takes $\approx$1 hour to train the whole 2-step model (including all pre-training steps) for the LOWO 2016--17 experiment. Predictions for a complete winter take only a couple of minutes.
\subsection{Ablation studies}
We ablate various design choices of our model. To quantify variations in performance, we measure the Mean Absolute Deviation (MAD) of the water fraction: we first compute absolute differences (AD) between the daily predictions and the ground truth water fractions, then average them to obtain the MAD.
\subsubsection{Effect of auxiliary loss terms in step 2}
\begin{figure}[!]
    \begin{tikzpicture}
        \centering
        \begin{axis}
        [
            width  = 1.0\linewidth,
            height = 4.5cm,
            major x tick style = transparent,
            ybar,
            bar width=7pt,
            ymajorgrids = true,
            ylabel = {Mean Absolute Difference ($\%$)},
            y label style={font=\scriptsize,align=center},
            y tick label style={font=\scriptsize,align=center},
            symbolic x coords={Sihl, Sils, Silvaplana, St.~Moritz},
            xtick = data,
            scaled y ticks = false,
            x tick label style={font=\scriptsize,text width=2.5cm,align=center},
            enlarge x limits=0.1,
            legend style={at={(0.5,-0.230)}, anchor=north,legend columns=-1},
            legend style={font=\scriptsize},
            ymax=20,
        ]
            \addplot[style={BLACK,fill=CYAN,mark=none}]
                coordinates {(Sihl,4.83) (Sils,9.38) (Silvaplana,13.16) (St.~Moritz,14.42)};
            \addplot[style={BLACK,fill=DARKSILVER,mark=none}]
                coordinates {(Sihl,9.45) (Sils,14.43) (Silvaplana,15.62) (St.~Moritz,16.49)};
            \addplot[style={BLACK,fill=LIGHTRED,mark=none}]
                coordinates {(Sihl,4.92) (Sils,9.34) (Silvaplana,13.01) (St.~Moritz,14.42)};
            \legend{$L_{reg}$, $L_{reg}-\beta L_{line}$, $L_{reg}-\gamma L_{idc}$}
        \end{axis}
    \end{tikzpicture}
    \caption{Bar graphs showing the effect of dropping auxiliary loss terms ($L_{line}$ and $L_{idc}$) in regression (step 2).}
    \label{fig:ablation:aux_loss}
\end{figure}
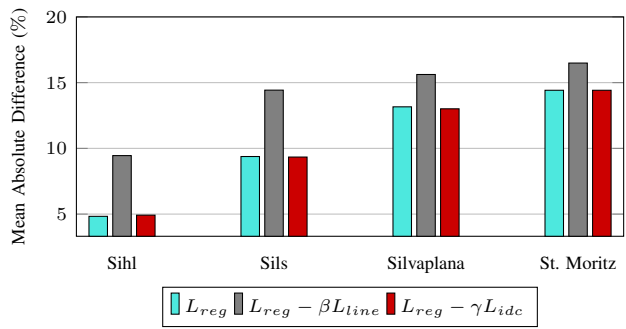

To study the effect of the auxiliary loss terms ($L_{line}$ and $L_{idc}$) in the regression step, we drop either of them from the regression loss $L_{reg}$ and retrain the model, with fixed initialisation to suppress stochastic fluctuations. The results are shown in Fig.~\ref{fig:ablation:aux_loss}. Dropping the \textit{line loss} ($L_{reg}-\beta L_{line}$) significantly increases the MAD. Dropping the \textit{intra-day coherence loss} ($L_{reg}-\gamma L_{idc}$) has a negligible effect. Apparently, intra-day coherence is already achieved without a dedicated loss term. We nevertheless recommend keeping $L_{idc}$: it has no adverse effects, and the finding that intra-day coherence is rarely violated is, for the time being, purely empirical and not backed up by a conceptual explanation. It may not generalise to other setups or regions of interest.
\subsubsection{Effect of window size in step 2}
\begin{figure}[!]
    \begin{tikzpicture}
        \centering
        \begin{axis}
        [
            width  = 1.0\linewidth,
            height = 4.5cm,
            major x tick style = transparent,
            ybar,
            bar width=7pt,
            ymajorgrids = true,
            ylabel = {Mean Absolute Difference ($\%$)},
            y label style={font=\scriptsize,align=center},
            y tick label style={font=\scriptsize,align=center},
            symbolic x coords={Sihl, Sils, Silvaplana, St.~Moritz},
            xtick = data,
            scaled y ticks = false,
            x tick label style={font=\scriptsize,text width=2.5cm,align=center},
            enlarge x limits=0.1,
            legend style={at={(0.5,-0.230)}, anchor=north,legend columns=-1},
            legend style={font=\scriptsize},
            ymin=4,
            ymax=20,
        ]
            \addplot[style={BLACK,fill=LIGHTRED,mark=none}]
                coordinates {(Sihl,7.156666667) (Sils,9.816666667) (Silvaplana,13.90333333) (St.~Moritz,14.34333333)};
            \addplot[style={BLACK,fill=CYAN,mark=none}]
                coordinates {(Sihl,5.686666667) (Sils,9.43) (Silvaplana,12.99) (St.~Moritz,14.11)};
            \addplot[style={BLACK,fill=LIGHTGREEN,mark=none}]
                coordinates {(Sihl,7.583333333) (Sils,12.01666667) (Silvaplana,13.45333333) (St.~Moritz,14.46333333)};
            \legend{$w=5$, $w=7$, $w=9$}
        \end{axis}
    \end{tikzpicture}
    \caption{Bar graphs showing the effect of window size ($w$) in regression (step 2).}
    \label{fig:ablation:win_size}
\end{figure}
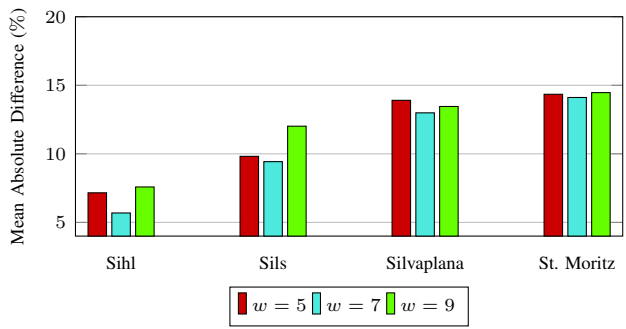

To investigate the effect of the temporal component in the regression head (\textit{window size} $w$), we report results with $w\in\{5,7,9\}$, see Fig.~\ref{fig:ablation:win_size}. For robustness, an ensemble over three independently initialised networks was used. 
Consistently across all four lakes, the lowest MAD is obtained with window size 7; while, overall, the method is not overly sensitive to this hyper-parameter. 
%
\section{Discussion and Conclusion}
We have developed a deep learning framework that learns a joint satellite embedding to fuse MODIS, VIIRS and S1-SAR satellite data into a task-specific, homogenised time-series. We have employed the proposed framework for lake ice monitoring, to estimate the freezing state of four Swiss lakes and to determine their ice-on and ice-off dates.
\par
In our experiments, the method has shown good generalisation across different winters and different lakes.
\par
Notably, the data we fuse for our application example not only come from two different imaging modalities but also differ in resolution by an order of magnitude. At least conceptually, the proposed multi-branch architecture is very generic. It should be straightforward to extend it to further satellite sensors, by pre-training the corresponding low-level branches and fine-tuning the shared block. 
\par
Although the proposed approach in many cases achieves promising results, there is still room for improvement. In particular, we have used only rough, approximate reference annotations for the transition days. Better ground truth will likely improve the performance of the regression. Moreover, we expect that adding optical satellite images with higher spatial resolution and not too long revisit times could improve the results. Given the statistical nature of our approach, it is also likely that a larger and more diverse training set would bring performance gains, but we leave this for future work, as it will require annotations for additional lakes and/or winters.

In our current setup, the network is trained in stages. Training all embedding branches and the regression head end-to-end would be an interesting and conceptually pleasing extension.
\par
For the regression, we compute the \textit{line loss} and the \textit{intra-day coherence loss} between the predictions, i.e., at the decision-level. Including coherence/continuity constraints already at the embedding level could potentially further improve the representation and maybe an interesting avenue for future work. An intriguing, but challenging direction would be to extend the fusion also to data from non-satellite sources (webcams, UAVs, etc.). 
\par
A caveat of our approach is that the learnt embedding at this stage is perhaps better described as \emph{equivariant}, in the sense of delivering comparable output for different image sources. But it is not truly invariant, given that the latent vectors to some degree still appear to cluster according to the underlying sensors.
\par
In terms of application, while our focus was on lake ice, the proposed data fusion methodology is generic and should be easily transferable to other geo-spatial data analysis and Earth science applications.
\par
Finally, at the methodological level, it may in the longer term be necessary to shift the focus to techniques that learn task-oriented data representations (i.e., embedding) with minimal supervision, such as unsupervised, weakly supervised or self-supervised learning. There exists a huge amount of unlabelled remotely sensed data, from various sources (spaceborne, airborne, webcams, amateur images, etc.), and labelling reference data for supervised learning is a critical bottleneck.


%

\section*{Acknowledgment}
The authors would like to thank Roberto Aguilar (Aerialytics, Costa Rica) for his support.
\appendices

\section{Details of target lakes}\label{app:A}
Properties of the four target lakes are displayed in Table~\ref{table:lakes}.
\begin{table}[h!]
\centering
\normalsize
\resizebox{0.45\textwidth}{!}{%
\begin{tabular}{lllll}
 \toprule
& \textbf{Sihl} & \textbf{Sils} & \textbf{Silvaplana} & \textbf{St.~Moritz}\\
 \midrule
\textit{Area~($km^2$)} & 11.3 & 4.1 & 2.7 & 0.78\\
\textit{Altitude~($m$)} & 889 & 1797 & 1791 & 1768 \\
\textit{Max.~depth~($m$)} & 23 & 71 &  77&   42\\
\textit{Avg.~depth~($m$)} & 17 & 35 &  48&   26\\
\textit{Volume~($Mm^3$)} & 96 & 137 &  140 &   20\\
 \bottomrule
\end{tabular}}
\caption{Physical properties of target lakes.}
\label{table:lakes}
\end{table} 

\section{Full winter time-series}
\label{appendix:B}
\begin{figure*}[h!]
    \centering
    \subfloat[\centering Lake Silvaplana]{{\includegraphics[width=.825\linewidth]{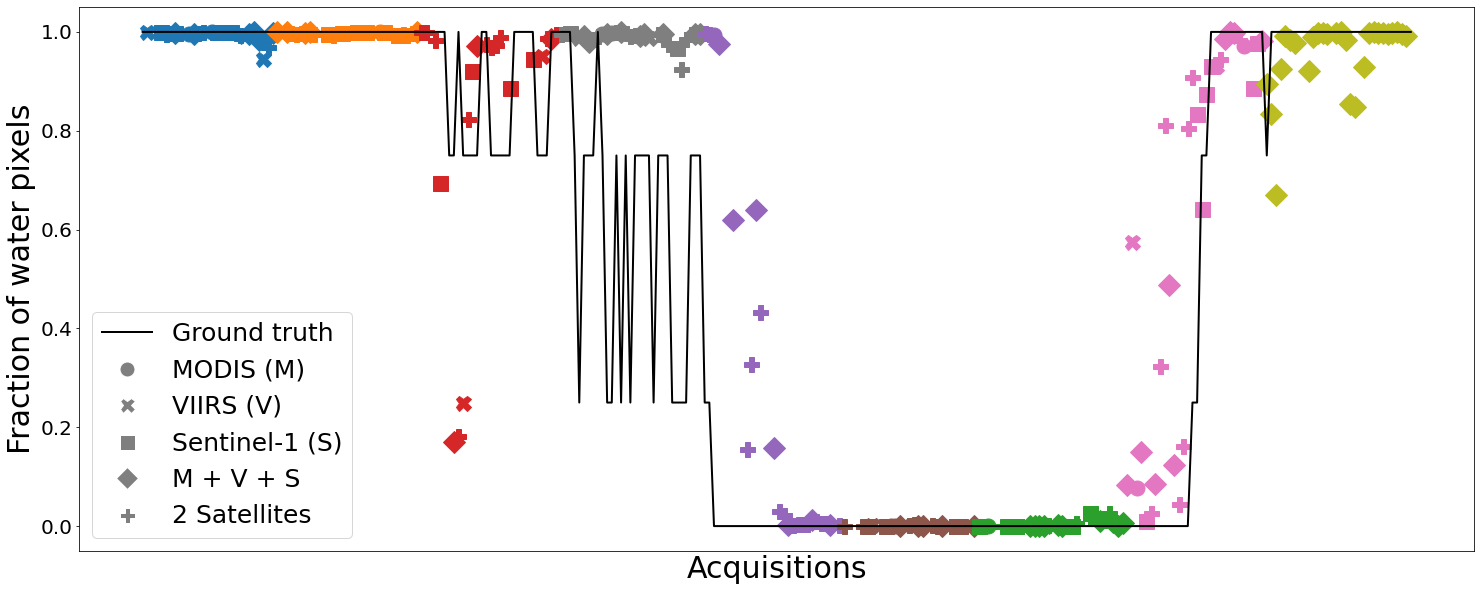}}}\\
    \subfloat[\centering Lake St.~Moritz]{{\includegraphics[width=.825\linewidth]{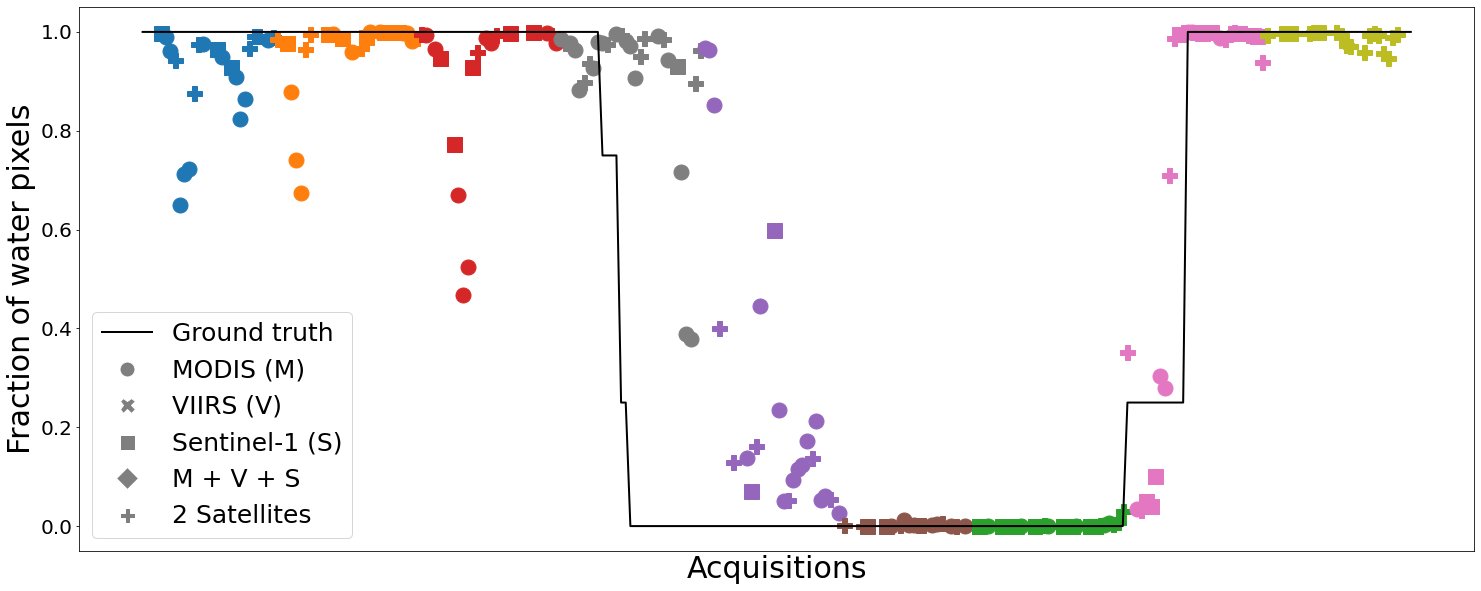} }}
    \caption{Time-series plots for lakes Silvaplana and St.~Moritz from winter 2016--17 using a model trained on all the data (all four lakes) from winter 2017--18. Predictions are separated by colours month-wise. Best if viewed on screen.}%
    \label{fig:time_series_2step_silvaplana_stmoritz}%
\end{figure*}

Full winter time-series results for lakes Silvaplana and St.~Moritz are displayed in Fig.~\ref{fig:time_series_2step_silvaplana_stmoritz}.
%
\section{Embedding visualisation}
\label{appendix:C}
\begin{figure*}[!]
    \centering
    \subfloat[Lake Silvaplana,  dimension=3]{\includegraphics[width=.425\linewidth]{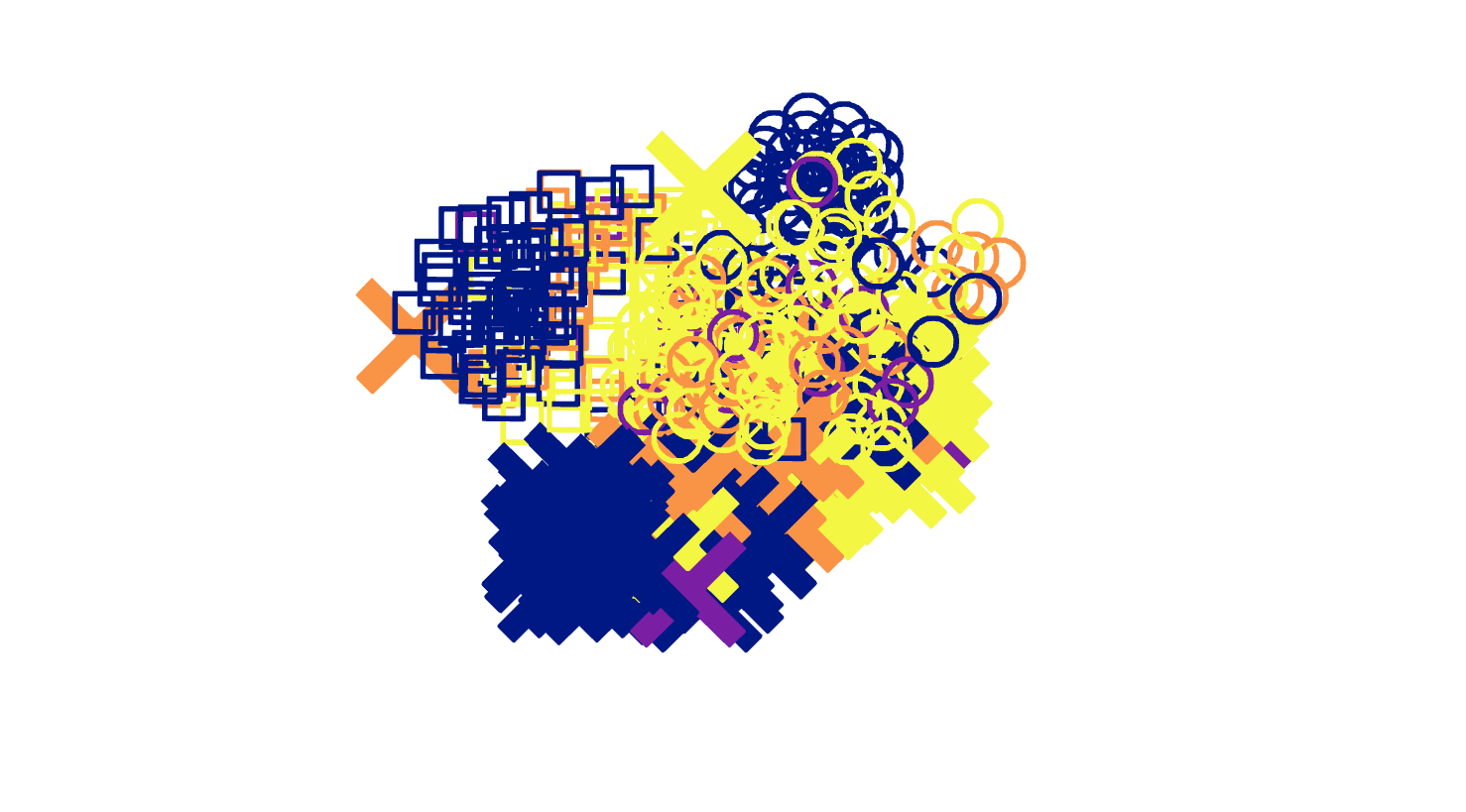}}
    \subfloat[Lake Silvaplana,  dimension=2]{\includegraphics[width=.425\linewidth]{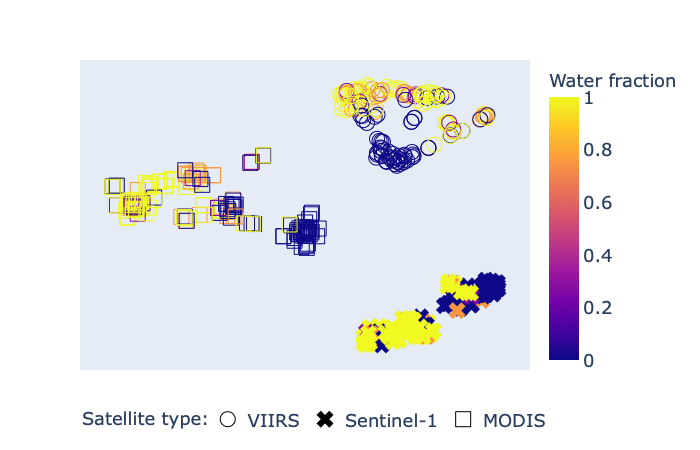}}\\
    \subfloat[Lake St.~Moritz,  dimension=3]{\includegraphics[width=.425\linewidth]{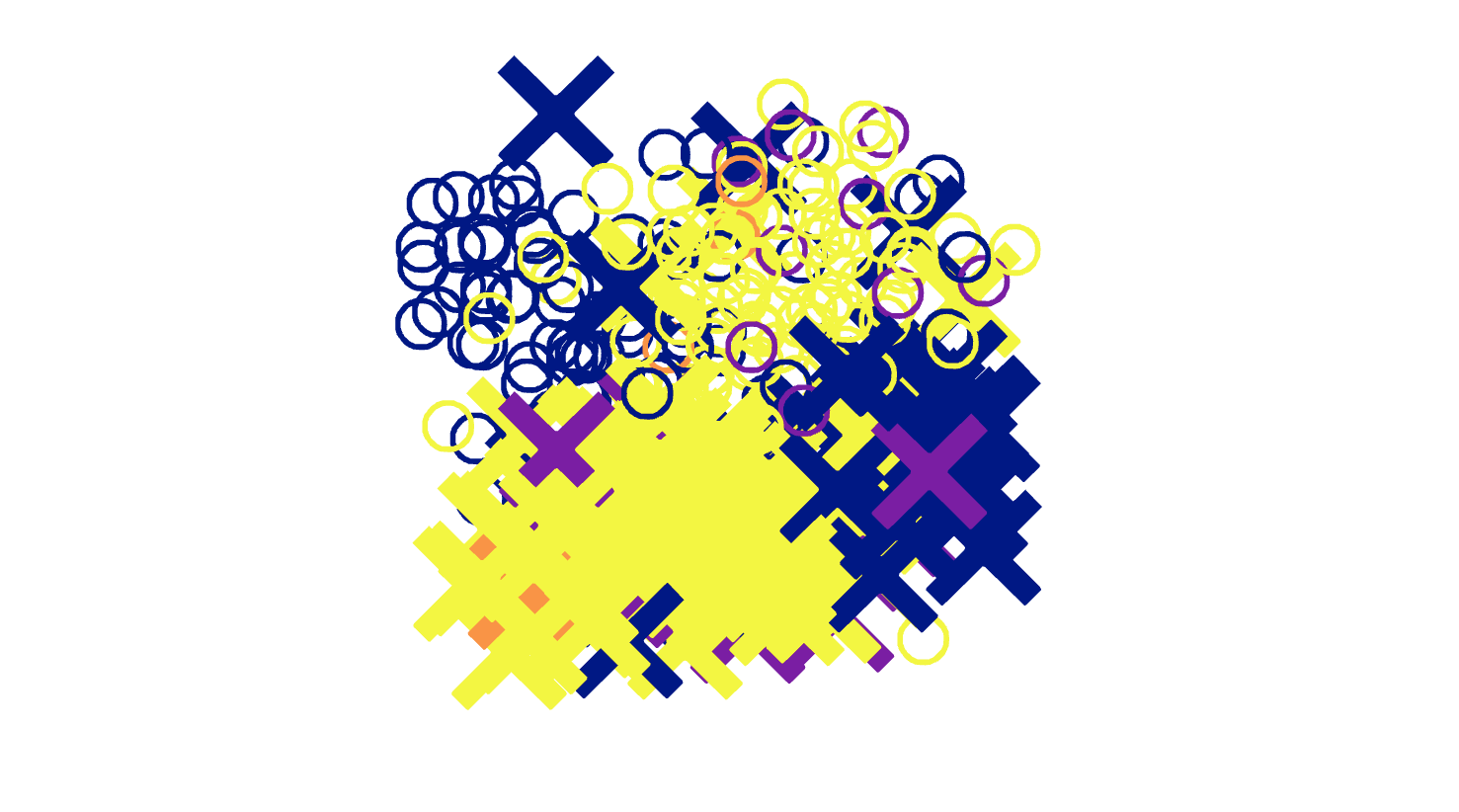}}
    \subfloat[Lake St.~Moritz,  dimension=2]{\includegraphics[width=.425\linewidth]{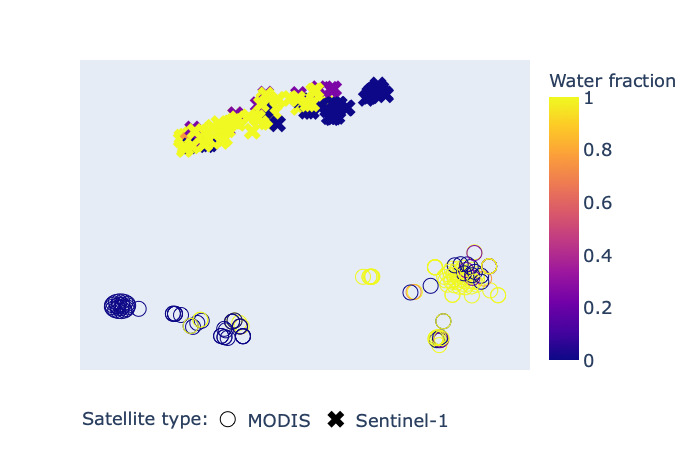}}\\
    \caption{t-SNE representation of the embedding learnt (lakes Silvaplana and St.~Moritz, winter 2016--17) using the proposed \textit{2-step} approach with a model trained on all the data from winter 2017--18. Water fraction refers to the fraction of non-frozen pixels. Different marker types are used to distinguish embeddings of different sensors. Best if viewed on screen.}%
    \label{fig:emb_vis_2}%
\end{figure*}

t-SNE representation for the learnt embedding for lakes Silvaplana and St.~Moritz are displayed in Fig~\ref{fig:emb_vis_2}.
%
\bibliographystyle{IEEEtran}

\bibliography{bibtex/bib/references}
%



%
%
\begin{IEEEbiography}[{\includegraphics[width=1.05in,trim=35mm 5mm 25mm 5mm, clip]{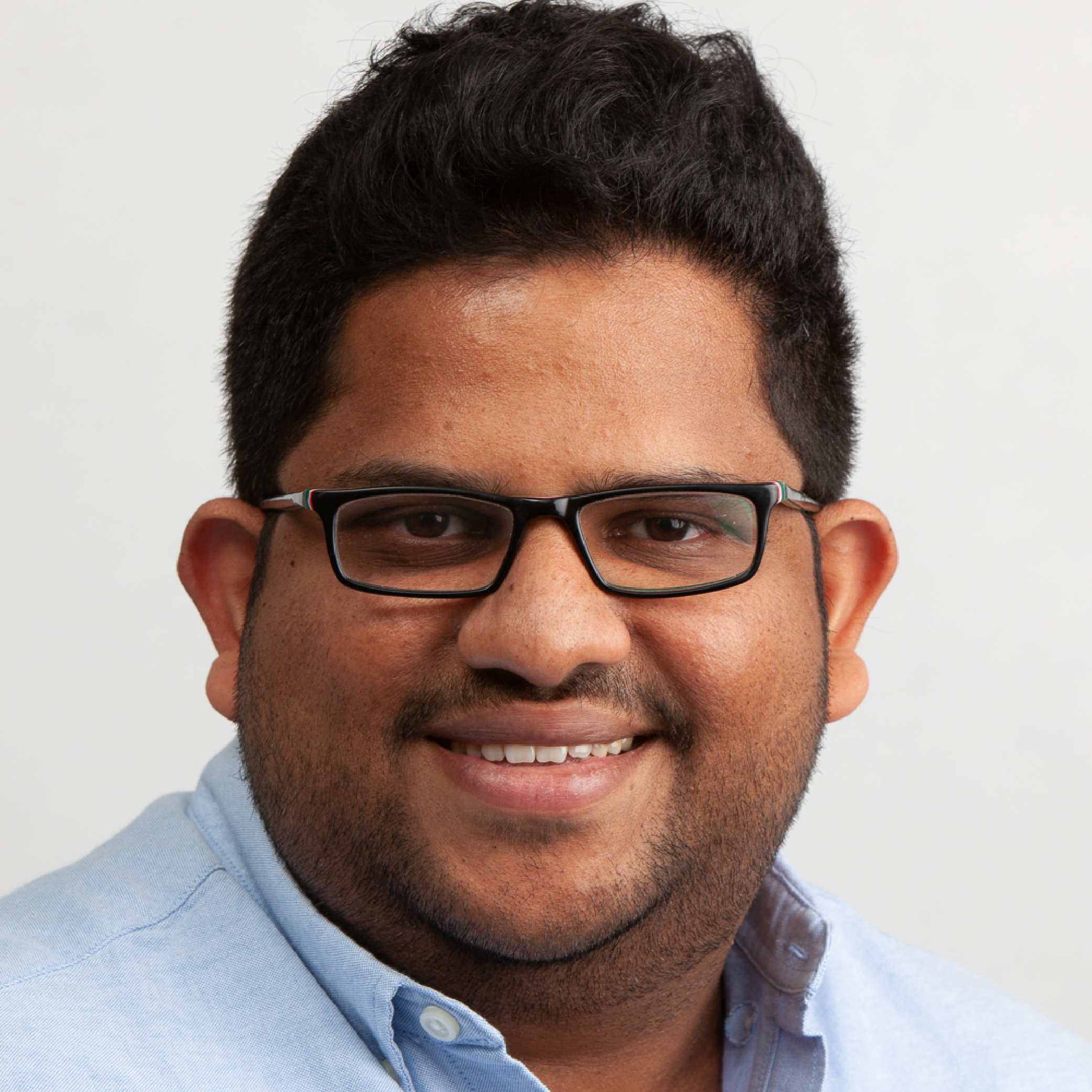}}]{Manu Tom}
received B.Tech. degree (Electronics and Communication Engineering, 2010) from the University of Kerala, Thiruvananthapuram, India. Then, he worked in the industry, followed by a research position at the Indian Institute of Science (Bangalore). Later, he received M.Sc. degree (Electrical Engineering, 2016) from the RWTH Aachen University, Germany, and Ph.D. degree (Geomatics Engineering, 2021) from ETH Zurich, Switzerland. At present, he is a post-doctoral researcher jointly at the Swiss Federal Institute of Aquatic Science and Technology (Eawag, Dübendorf) and the University of Zurich, Switzerland. His research interests include remote sensing of the environment and natural hazards, computer vision, geospatial data analysis, and satellite data fusion.
\end{IEEEbiography}
%
\begin{IEEEbiography}[{\includegraphics[width=1.05in,trim=30mm 5mm 25mm 5mm, clip]{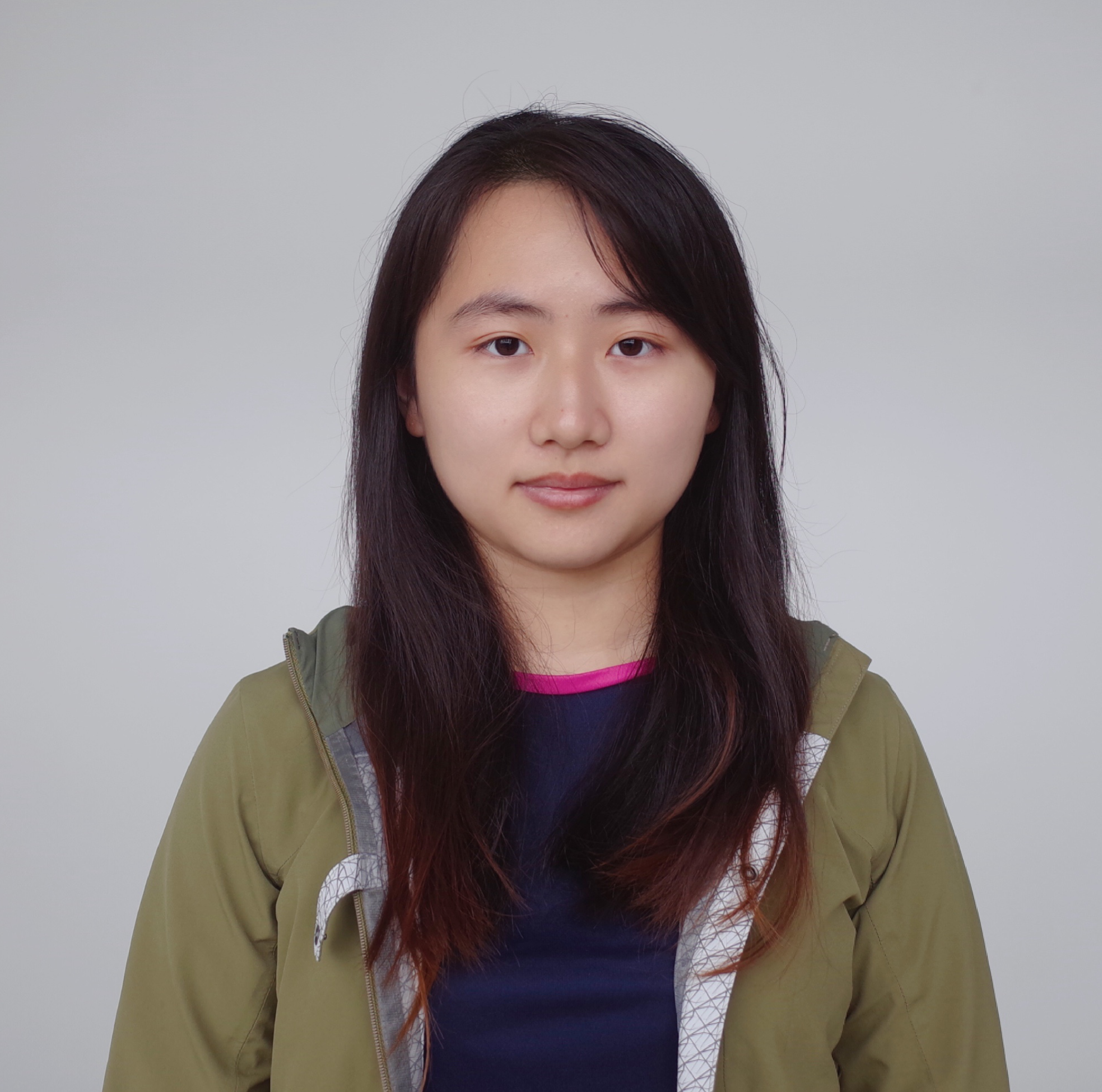}}]{Yuchang Jiang}
received B.S. in Geomatics from The Hong Kong Polytechnic University, Hong Kong in 2019 and  M.S. in Geomatics from the ETH Zurich, Switzerland in 2021. Her research interests include computer vision, remote sensing and machine learning.
\end{IEEEbiography}
%
%
\begin{IEEEbiography}[{\includegraphics[width=1.075in,clip, height=1.1in]{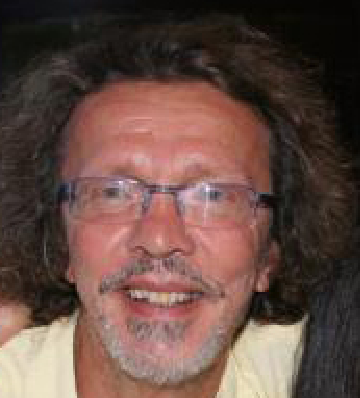}}]{Emmanuel Baltsavias} received his Dipl. Ing. in Surveying and Rural Engineering at the National Technical University of Athens in 1981, MSc in Digital Photogrammetry in 1984 from Ohio State University and his PhD. on image matching in 1991 from ETH Zurich. He is currently a Senior Scientist and Lecturer at the Institute of Geodesy and Photogrammetry, ETH Zurich. He was elected as ISPRS Fellow in 2010 and also served as the Vice President of ISPRS (2004-2008). His research interests include automated extraction and classification of objects, data and information fusion, processing of high spatial resolution optical sensors, investigations of airborne digital photogrammetric cameras, applications in forestry, natural hazards and cultural heritage.
\end{IEEEbiography}
\begin{IEEEbiography}[{\includegraphics[width=1.05in,trim=27mm 5mm 25mm 5mm, clip]{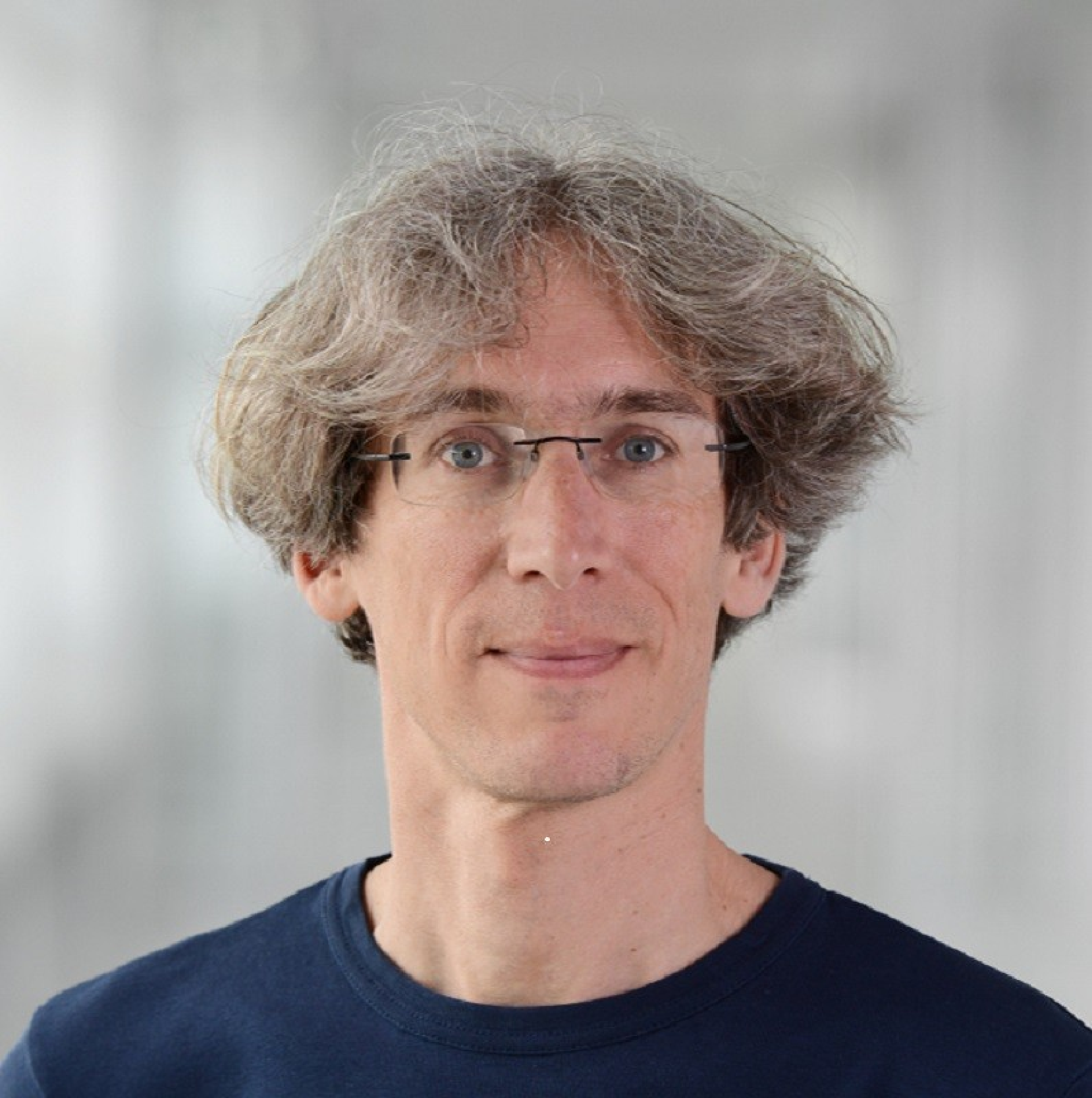}}]{Konrad Schindler}
(M’05–SM’12) received the Diplomingenieur (M.Tech.) degree from Vienna University of Technology, Vienna, Austria, in 1999, and the Ph.D. degree from Graz University of Technology, Graz, Austria, in 2003. He was a Photogrammetric Engineer in the private industry and held research positions at Graz University of Technology, Monash University, Melbourne, VIC, Australia, and ETH Zurich, Switzerland. He was an Assistant Professor of Image Understanding with TU Darmstadt, Darmstadt, Germany, in 2009. Since 2010, he has been a Tenured Professor of Photogrammetry and Remote Sensing with ETH Zurich. His research interests include computer vision, photogrammetry, and remote sensing.
\end{IEEEbiography}
\end{document}